\documentclass[10pt,twocolumn,letterpaper,table]{article}

\usepackage[pagenumbers]{cvpr} %

\usepackage{amsmath,amsfonts,bm,amssymb,mathtools,amsthm}
\usepackage{color,xspace}
\usepackage{booktabs}
\usepackage{thm-restate}

\def\eqref#1{Eq.~(\ref{#1})}

\def\1{\bm{1}}

\def\rvepsilon{{\mathbf{\epsilon}}}
\def\rvtheta{{\mathbf{\theta}}}

\def\rvc{{\mathbf{c}}}

\def\rvu{{\mathbf{i}}}

\def\rvu{{\mathbf{u}}}

\def\rvx{{\mathbf{x}}}

\def\rmK{{\mathbf{K}}}

\def\rmQ{{\mathbf{Q}}}

\def\rmV{{\mathbf{V}}}
\def\rmW{{\mathbf{W}}}

\def\rmZ{{\mathbf{Z}}}

\DeclareMathAlphabet{\mathsfit}{\encodingdefault}{\sfdefault}{m}{sl}
\SetMathAlphabet{\mathsfit}{bold}{\encodingdefault}{\sfdefault}{bx}{n}

\newcommand{\E}{\mathbb{E}}

\newcommand{\KL}{\mathbb{D}_{\mathrm{KL}}}

\def\rvxw{\rvx^+}
\def\rvxl{\rvx^-}
\def\rvepsilonw{\rvepsilon^+}
\def\rvepsilonl{\rvepsilon^-}

\def\p{p_\rvtheta}
\def\q{p_\text{ref}}

\newcommand{\diffdpo}{Diffusion-DPO\xspace}
\newcommand{\model}{PPD\xspace}

\usepackage{multirow}

\usepackage[export]{adjustbox}

\usepackage[column=0]{cellspace}
\definecolor{blue1}{RGB}{0, 161, 255}
\definecolor{green1}{RGB}{97, 216, 54}
\definecolor{grey1}{RGB}{146, 146, 146}

\def\maketitlesupplementaryonecolumn
   {
   \newpage
       \onecolumn{
        \centering
        \Large
        \textbf{\thetitle}\\
        \vspace{0.5em}Supplementary Material \\
        \vspace{1.0em}
       }
   }

\definecolor{cvprblue}{rgb}{0.21,0.49,0.74}
\usepackage[pagebackref,breaklinks,colorlinks,allcolors=cvprblue]{hyperref}
\usepackage[linesnumbered,lined,boxed,commentsnumbered,ruled,vlined, noend]{algorithm2e}
\usepackage{amsmath}
\usepackage{tabularx}

\title{
Personalized Preference Fine-tuning of Diffusion Models
}

\makeatletter
\renewcommand{\and}{\unskip ~~~~~~~}
\makeatother

\author{Meihua Dang$^{1*}$ 
\and Anikait Singh$^{1*}$ 
\and Linqi Zhou$^{1,2}$
\and Stefano Ermon$^{1}$ 
\and Jiaming Song$^{2}$ \\
$^1$Stanford University, $^2$Luma AI\\
{\tt\small \{mhdang,anikait,lzhou907,ermon\}@cs.stanford.edu,} 
{\tt\small jiaming@lumalabs.ai} 
}

\begin{document}
\maketitle

\begin{abstract}

RLHF techniques like DPO can significantly improve the generation quality of text-to-image diffusion models. However, these methods optimize for a \textbf{single reward} that aligns model generation with population-level preferences, neglecting the nuances of individual users' beliefs or values. This lack of personalization limits the efficacy of these models. To bridge this gap, we introduce \model, a \textbf{multi-reward} optimization objective that aligns diffusion models with \textbf{personalized} preferences. With \model, a diffusion model learns the individual preferences of a population of users in a few-shot way, enabling generalization to unseen users. Specifically, our approach (1) leverages a vision-language model (VLM) to extract personal preference embeddings from a small set of pairwise preference examples, and then (2) incorporates the embeddings into diffusion models through cross attention. Conditioning on user embeddings, the text-to-image models are fine-tuned with the DPO objective, simultaneously optimizing for alignment with the preferences of multiple users. Empirical results demonstrate that our method effectively optimizes for multiple reward functions and can interpolate between them during inference. In real-world user scenarios, with as few as four preference examples from a new user, our approach achieves an average win rate of {76\%} over Stable Cascade, generating images that more accurately reflect specific user preferences.

\end{abstract}

\let\thefootnote\relax\footnotetext{$^*$Equal contributions.}
\let\thefootnote\relax\footnotetext{Work done during MD's internship at Luma AI.}
\section{Introduction}
\label{sec:intro}

Diffusion models~\cite{ddpm,sohl2015deep,song2019generative,song2021scorebased} have achieved great success in text-to-images generation tasks, as demonstrated by models such as DALL-E~2~\cite{ramesh2022hierarchical}, Imagen~\cite{Saharia2022photorealistic}, Stable Diffusion~\cite{rombach2022sd15,podell2024sdxl}, and Stable Cascade~\cite{pernias2024wrstchen}. These large-scale models, often trained through supervised fine-tuning on text-image datasets containing high-quality aesthetic images such as LAION-5B~\cite{Schuhmann22laion}, demonstrate impressive image quality and a remarkable visual understanding of text prompts. 
Recent advances focus on fine-tuning diffusion models to better align with human preferences. Methods such as DDPO~\cite{black2024ddpo} and DPOK~\cite{fan2024dpok} employ reinforcement learning from human feedback (RLHF) techniques to optimize human-alignment reward functions. Alternatively, \diffdpo~\cite{wallace2024diffusion} utilize direct preference optimization (DPO)~\cite{rafailov2024direct} to directly optimizes for human preferences using an offline ranking dataset. SD3~\cite{esser2024sd3} shows that fine-tuning diffusion models with human feedback techniques can significantly enhance overall image quality.
 
Despite these promising results, \diffdpo aligns a model with a single reward that represents population-level human preferences. These fine-tuning techniques are thus insufficient for achieving personalized preference alignment. As studied in domains such as language~\cite{2023arXiv230317548S,2024arXiv240615951F,2024arXiv240205070S}, individual users have unique preferences—for instance, some may favor images with brighter colors, while others might prefer images with centralized foregrounds. Fine-tuning a separate model for each person based on their preference data is impractical and does not scale well, limiting the feasibility of personalized alignment. Additionally, this may limit cross-user generalization.

A complementary research direction introduces additional control mechanisms into diffusion models to enable customization and personalization. Approaches such as IP-Adapter~\cite{ye2023ipadapter} and CtrlNet~\cite{zhang2023ctrlnet} allow for controlled image generation by using an additional conditioning image that encodes personalization features. However, these methods are constrained by their reliance on single-image inputs and do not directly learn or adapt to human preferences.

To address these limitations, we propose \model, a general framework that incorporates personalization into a diffusion model preference alignment through multi-reward optimization. Our approach learns a diverse set of human preferences within a single model and generalizes effectively to new users. Personalization in our approach requires minimal additional data, as we label samples in the preference dataset with a unique identifier for each user. This annotation is inexpensive to obtain and commonly included in datasets such as Pick-a-Pic~\cite{pickscore}. We leverage vision-language models (VLMs) such as LLaVA-OneVision~\cite{li2024llava} to extract user preference embeddings from few-shot pairwise preference examples for each user. We show that these few-shot text embeddings from preference pairs serve as effective features to represent reward functions under the Bradley-Terry model~\cite{bradley1952rank} and thus are suitable for conditioning the diffusion model. Next, adopting Stable Cascade~\cite{pernias2024wrstchen} as our base model, we condition the diffusion model on these user embeddings through additional cross-attention layers and fine-tune it with a variant of the \diffdpo~\cite{wallace2024diffusion} objective for preference alignment.

Our empirical evaluations demonstrate that \model can effectively optimize multiple rewards including CLIP~\cite{laionclip,clip}, Aesthetic~\cite{schuhmann22aes}, and HPS~\cite{wu2023hpsv2}, significantly outperforming baselines. It also interpolates smoothly between different rewards during inference. Additionally, in real-world user scenarios, \model aligns more closely with individual user preferences on the dataset Pick-a-Pic~\cite{pickscore}. Notably, with only a few-shot ranking examples for a new user,  \model generate images that align closely with that user’s specific preference and outperforms prior approaches such as Stable Cascade, achieving an average win rate of {76\%}, as evaluated by GPT4o~\cite{OpenAI2023GPT4TR}.

\begin{figure*}
    \centering
    \includegraphics[page=3,width=\linewidth]{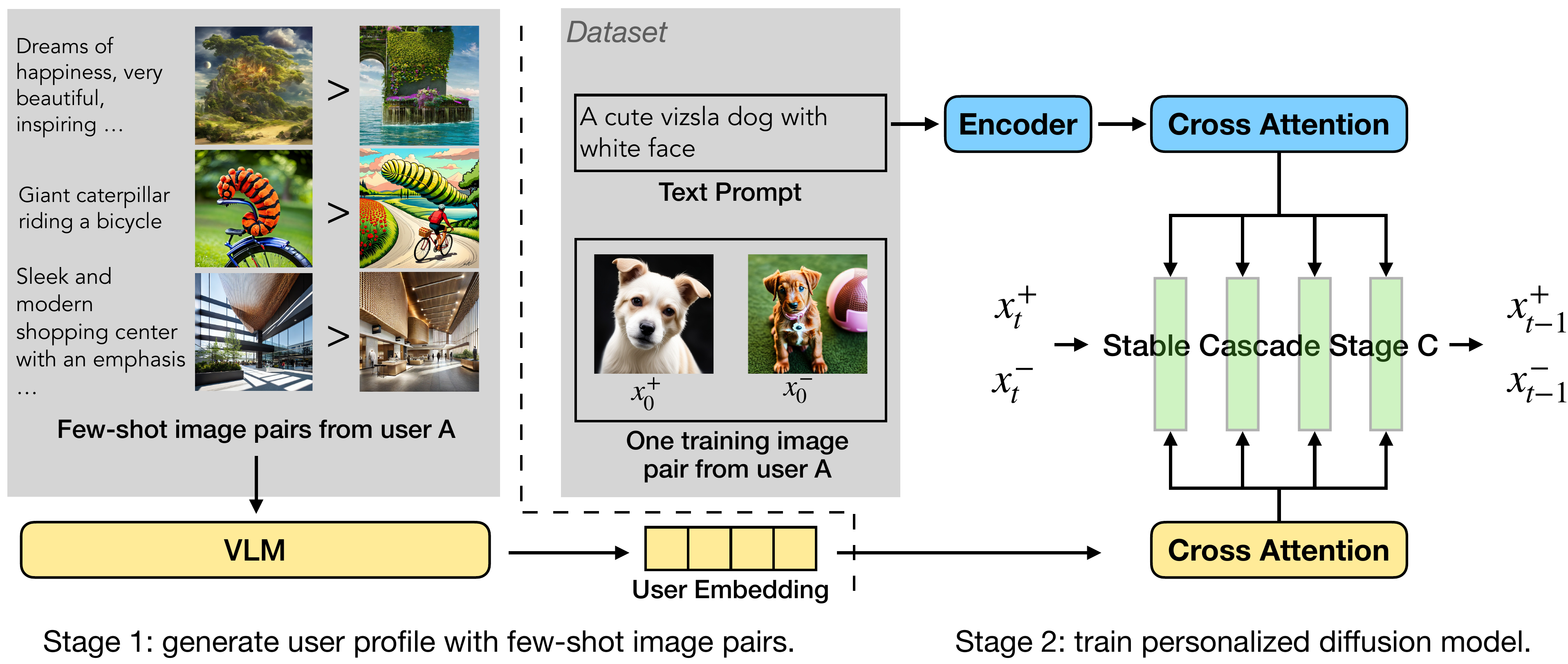}
    \caption{\textbf{The overall architecture of \model.} In Stage 1, user embedding are generated with few-shot preference examples using a VLM. In Stage 2, we fine-tune diffusion models on the preference datasets with the user embedding as conditioning added to cross-attention.}
    \label{fig:pipeline}
\end{figure*}

\section{Related Work}
\label{sec:related}
\paragraph{Aligning Diffusion Models} Aligning text-to-image diffusion models has been extensively studied. Some methods optimize with online RL using a Bradley Terry reward function~\cite{black2024ddpo, fan2024dpok, zhang2024large}. Others fine-tune diffusion models to increase the reward of generated images by gradient back-propagation on the reward function~\cite{clark2023directly, xu2024imagereward, prabhudesai2023alignprop}. 
Recently proposed approaches~\cite{wallace2024diffusion, yang2024d3po} leverage the duality of the reward learning and policy optimization from DPO~\citep{rafailov2024direct} to eliminate the need of on-policy samples and only utilize offline human preference data. Alternatively, DiffusionKTO~\cite{li2024aligning} replaces DPO with KTO~\cite{ethayarajh2024kto}, eliminating the need for paired data. 
Additionally, some methods modify the training data distribution on visually appealing and textually cohered data~\cite{wu2023human,lee2023aligning,dai23emu}, while others improve text accuracy by re-captioning pre-collected web images~\cite{betker2023improving,segalis2023picture}. 

\paragraph{Controllable Generation for Diffusion Models}
To address the lack of fine-grained control in image synthesis, various methods add conditional controls to text-to-image diffusion models. CtrlNet~\cite{zhang2023ctrlnet} and T2I-Adapter~\cite{mou2024t2i} introduce additional UNet modules that encode spatial information, such as edges and human poses, derived from conditioning images. IP-Adapter~\cite{ye2023ipadapter} introduces additional cross-attention layers to incorporate high-level semantic information from reference images. 

\paragraph{LLM Alignment} 
The alignment of large language models (LLMs) using RLHF objectives has been extensively explored \cite{zheng2023secrets, bai2022training, rafailov2024direct}, demonstrating effective instruction-following and alignment to population values. However, a key challenge remains in representing the diversity of human preferences~\cite{2023arXiv230317548S}, as these models often capture an averaged population preference rather than accounting for individual differences, leading to biases and reduced alignment with underrepresented communities. \emph{Pluralistic Alignment} \cite{2024arXiv240615951F, 2024arXiv240205070S} addresses this issue by modeling a distribution of individuals, each with distinct values and perspectives, with methods like distributional and overton alignment enhancing calibration across diverse user groups. Several works have investigated LLM personalization through the reparameterization of reward models \cite{poddar2024personalizing, yu2024few, li2024personalized}, enabling a more tailored user experience. However, these works leverage synthetic users and study a limited number of users, particularly in text and robotic control domains. 

\section{Preliminary}
\label{sec:preliminary}
We introduce the preliminaries behind diffusion models, DPO and its adaption to diffusion models.

\paragraph{Diffusion Models}
Diffusion models are generative models consisting two processes: a forward (diffusion) process and a backward (denoising) process~\cite{ddpm,sohl2015deep,song2019generative,song2021scorebased}. In the forward process, Gaussian noise  $\rvepsilon\!\sim\!\mathcal{N}(\mathbf{0}, \mathbf{I})$ is progressively added to a data distribution $P_\text{data}(\rvx_0)$ over $T$ steps, guided by noise scheduling functions $\alpha_t$ and $\sigma_t$. At timestep $t\in[0,T]$, noisy images $\rvx_t$ are generated via $\rvx_t=\alpha_t \rvx_0 + \sigma_t \rvepsilon$. 
In the backward process, samples are generated from Gaussian noise with a learnable model, which can also be conditioned on additional inputs, such as text in the case of text-to-image diffusion models. Typically, diffusion models -- denoted as $\rvepsilon_\rvtheta$ -- are trained to predict the Gaussian noise $\rvepsilon$ from noisy images $\rvx_t$ at timestep $t$ with conditioning $\rvc$, using a denoising objective:
\begin{equation}
    L_{\text{DM}} = \E_{\rvx_0, \rvepsilon, t}\left[\omega(\lambda_t)\|\rvepsilon-\rvepsilon_\rvtheta(\rvx_t,\rvc,t)\|_2^2\right],
    \label{eq:dm}
\end{equation}
where $\lambda_t=\log\left[\alpha_t^2/\sigma_t^2\right]$ is a signal-to-noise ratio, $\omega(\lambda_t)$ is a pre-specified weighting function~\cite{vaediffusion}. Once the model $\rvepsilon_\rvtheta$ is trained, images can be generated from random noise using accelerated sampling methods such as DDIM~\cite{song2021denoising}.

\paragraph{Direct Preference Optimization (DPO)} DPO~\cite{rafailov2024direct} is a fine-tuning paradigm that aligns autoregressive large language models (LLMs) with human preferences without relying on traditional reinforcement learning techniques. Unlike Reinforcement Learning from Human Feedback (RLHF), which involves training a reward model and then fine-tuning the LLM using reinforcement learning, DPO directly optimizes the model based on human preference data.  Given a dataset containing ranking pairs $(\rvc,\rvxw_0,\rvxl_0)$ where $\rvc$ being the prompt and $\rvxw_0,\rvxl_0$ being the preferred and dispreferred responses labeled by human, the training objective optimizes a simple loss under which we can obtain an optimal policy:
\begin{align}
\begin{split}
    L_{\text{DPO}}(\rvtheta)=&-
    \E_{\rvc,\rvxw_0,\rvxl_0}\left[
    \log\sigma \left( \beta \Delta\right) \right], \text{where} \\ 
    \Delta=&
    \left(\log \frac{\p(\rvxw_0|\rvc)}{\q(\rvxw_0|\rvc)}- \log \frac{\p(\rvxl_0|\rvc)}{\q(\rvxl_0|\rvc)}\right),
\end{split}
\label{eq:dpo}
\end{align}
where $\sigma$ is the sigmoid function and $\beta$ is a hyperparameter determining how much $\p(\rvx_0|\rvc)$ 
may deviate from a reference distribution 
$\q(\rvx_0|\rvc)$. 
Intuitively, the objective aims to increase the likelihoods of preferred responses while decreasing that of dispreferred ones.

\paragraph{Diffusion DPO}
The DPO objective in~\cref{eq:dpo} requires optimization of log-likelihoods $\log {\p(\rvxw_0|\rvc)}$, which is analytically intractable for diffusion models. \diffdpo ~\cite{wallace2024diffusion} leverages the fact that the (appropriately weighted) training objective of diffusion models as in~\cref{eq:dm} can be viewed as the evidence lower bound (ELBO) of their log-likelihoods. It derives an upper bound to DPO for diffusion models to align them with human preferences:
\begin{align}
    &L_\text{Diff-DPO}(\rvtheta)
    = - \E_{\rvc,\rvxw_0, \rvxl_0}\left[ 
    \log\sigma \left(-\beta T \omega(\lambda_t) \Delta \right)\right], \text{where~}\nonumber \\
    &\Delta  = \| \rvepsilonw -\rvepsilon_\theta(\rvxw_t,\rvc,t)\|^2_2 - \|\rvepsilonw - \rvepsilon_\text{ref}(\rvxw_t,\rvc,t)\|^2_2 \nonumber\\
     &\hphantom{\Delta}
 -\left( \| \rvepsilonl \!-\rvepsilon_\theta(\rvxl_{t}\!,\rvc,t)\|^2_2 - \|\rvepsilonl\! - \rvepsilon_\text{ref}(\rvxl_{t}\!,\rvc,t)\|^2_2\right),
\label{eq:diff-dpo}
\end{align}
where $\rvc$ is the caption; $\rvxw_0$ and $\rvxl_0$ are the preferred and dispreferred images; $\rvxw_t\!=\!\alpha_t \rvxw_0 + \sigma_t \rvepsilonw$ and $\rvxl_t\!=\!\alpha_t \rvxl_0 + \sigma_t \rvepsilonl$ are noisy images  obtained from forward process with Gaussian noise $\rvepsilonw$ and $\rvepsilonl$ at timestep $t$. This objective computes 4 terms as in~\cref{eq:dm}: the denoising loss of training and preference distributions $\rvepsilon_\theta$ and $\rvepsilon_\text{ref}$ on preferred and dispreferred noisy images $\rvxw_t$ and $\rvxl_t$ respectively. Intuitively, it aims to decrease the loss of preferred images while increasing that of dispreferred ones.

\section{Method}
\label{sec:method}

Despite impressive image quality and alignment control achieved in prior works, a key limitation among works that use DPO for diffusion models is that it assumes a uniform preference among all users for generated images.
In practice, this assumption often fails, as individuals may have diverse preferences related to style, color, lighting, and other visual attributes. While such preferences can be easily expressed by human through pairwise comparisons, it is challenging to summarize them accurately through text or single-image prompts, due to the inaccuracy of natural language and the limitations inherent in single-prompt formats.

Motivated by this problem, we propose \model, a method that aligns diffusion models with individual-level human preferences.
We first formulate the problem of personalization and define a new DPO variant for this task (\cref{sec:method-loss}), and then show that individual-level preference learning can be achieved using few-shot comparison examples using a VLM~(\cref{sec:method-vlm}), finally we introduce a parameterization of denoiser that can takes advantage of the VLM embeddings~(\cref{sec:method-ipadapter}). 
The overall architecture is in \cref{fig:pipeline}.
\subsection{Problem Setup}
\label{sec:method-loss}
We consider a dataset containing $(\rvc,\rvxw_0,\rvxl_0,\rvu)$ where $\rvc$ is the text prompt, $\rvxw_0,\rvxl_0$ are preferred and dispreferred generated images and $\rvu$ represents the user labeling the preference. User features $\rvu$ can be high-dimensional including information like age or gender, but often  $\rvu$  consists only of scalar user IDs $\rvu\!=\![u]$. We define $r(\rvc,\rvx_0, \rvu)$ as the reward on image $\rvx_0$ for user $\rvu$ given prompt $\rvc$. We would like to fine-tune a text-to-image models $p_{\theta_i}(\rvx_{0}|\rvc)$ for individual user $\rvu_i$ such that reward is maximized for this user while keeping close to a reference model $p_{\text{ref}}(\rvx_0|\rvc)$ in terms of KL-divergence as regularization :
\begin{align}
\begin{split}
    \max_{p_{\theta_i}} &\E_{\rvc, \rvx_0} \left[r(\rvc,\rvx_0, \rvu_i)\right]\\
    &-\beta \KL \left[p_{\theta_i}(\rvx_{0}|\rvc)\|p_{\text{ref}}(\rvx_0|\rvc)\right],
\end{split}
    \label{eq:dpo-objective-R}
\end{align}
where $\beta$ is a parameter controlling how much $p_{\theta_i}(\rvx_{0}|\rvc)$ deviates from $p_{\text{ref}}(\rvx_0|\rvc)$.
Following \citep{rafailov2024direct}, the optimal $p_{\theta_i}$ for user $u_i$ takes the form
\begin{equation}
    p_{\theta_i}(\rvx_0|\rvc) \propto \q(\rvx_0|\rvc)\exp \left(r(\rvc,\rvx_0, \rvu_i)/\beta\right).
\end{equation}
Suppose model $p_\theta$ has sufficient capacity, we can represent $p_{\theta_i}(\rvx_0|\rvc) $ as $p_{\theta}(\rvx_0|\rvc, \rvu_i) $, where  $\rvu_i$ serves as as a conditioning. This allows us to jointly optimize a single model $p_{\theta}(\rvx_0|\rvc, \rvu)$ for all users. Following \cite{wallace2024diffusion}, we have the following objective to learn a personalized diffusion models:
\begin{align}
    &L_\text{\model}(\rvtheta)
    = - \E_{\rvc,\rvxw_0, \rvxl_0,\textcolor{red}{\rvu}}\left[
    \log\sigma \left(-\beta T \omega(\lambda_t)\Delta \right)\right], \text{where}\nonumber \\
    &\Delta= \| \rvepsilonw -\rvepsilon_\theta(\rvxw_t,\rvc,\textcolor{red}{\rvu},t)\|^2_2 - \|\rvepsilonw - \rvepsilon_\text{ref}(\rvxw_t,\rvc,t)\|^2_2 \nonumber\\
    &-\left( \| \rvepsilonl \!-\rvepsilon_\theta(\rvxl_{t}\!,\rvc,\textcolor{red}{\rvu},t)\|^2_2 \!-\! \|\rvepsilonl\! - \rvepsilon_\text{ref}(\rvxl_{t}\!,\rvc,t)\|^2_2 \right).\!\label{eq:diff-dpo-u}
\end{align}
This objective is similar to \cref{eq:diff-dpo} but with the addition of $\rvu$, representing extra features conditioned on for each user. But what user features should be conditioned on?
\subsection{Generating User Features from a VLM with In-Context Examples}
\label{sec:method-vlm}
\paragraph{Discussion on User Features in Preference Datasets} One fundamental challenge in personalized image generation lies in the effective parameterization of the user representation, denoted as $\rvu$, for each individual. This parameterization is critical because it directly influences the model's capacity to generate outputs tailored to specific user preferences. Existing image preference datasets, such as Pick-a-Pic~\citep{pickscore}, typically lack comprehensive user information beyond basic identifiers, often focusing on generic preference data rather than capturing the nuanced individual preferences necessary for effective personalization. 

Even in preference datasets that include user information (e.g PRISM~\citep{2024arXiv240416019K}), this information is frequently obtained through self-reported surveys, which can lead to incomplete or biased user profiles. Discrepancies often arise between stated preferences and actual behavior; users may not accurately articulate their likes and dislikes, or they may respond in a manner that they perceive as socially desirable rather than truthful. Such inconsistencies present significant challenges when attempting to model and predict user preferences accurately.

To address these issues, we adopt a simplified approach by assuming access solely to a unique user ID for each individual in the dataset. This user ID serves as a key to differentiate users within the labeled preference dataset, represented as $\mathcal{D}_{\text{pref}} = \{(\mathbf{c}, \mathbf{x}_0^+, \mathbf{x}_0^-, \mathbf{s})^{(i)}\}$,  where $\mathbf{s}$ denotes a unique user identifier. 

\begin{figure}
    \centering
    \includegraphics[width=0.9\linewidth]{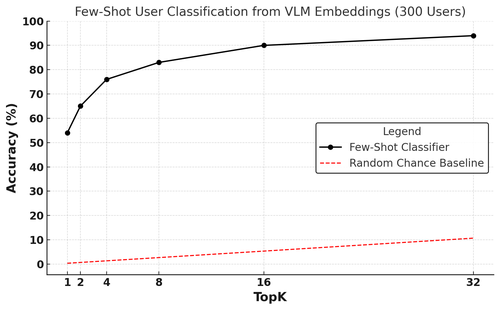}
    \caption{\textbf{Top-K accuracy of the User Classification}. We fine-tune a user-classifier from the frozen embeddings from the VLM on few-shot preference examples for 300 users. This classifier significantly outperforms a random chance baseline.}
    \label{fig:user-class}
\end{figure}

\paragraph{Conditioning on a Unique User Identifier} The primary challenge, then, is to leverage this user ID $i$ to construct a meaningful user embedding that can effectively guide the model's predictions. A naïve approach might involve representing the embedding $\rvu$ using a one-hot encoding of the user IDs, which can be formally defined by a function $f: \mathbf{S} \mapsto U$, establishing a bijective mapping from the set of user IDs $\mathbf{s} \in \{1, 2, \ldots, |D|\}$ to the space of one-hot encoded vectors $\rvu_i \in \{0,1\}^{|D|}$, where $|D|$ represents the total number of unique users in the dataset. While this method uniquely encodes each user, it suffers from a critical limitation: the inability to generalize to new, unseen users, as the embedding space is inherently tied to the users present in the training data. 

To enhance generalization and adaptability, we reconsider the parameterization of $u_i$ in the context of learning the distribution of reward functions, denoted as $r(\rvc, \rvx_0, \rvu_i)$. Here, a more flexible and robust parameterization leverages few-shot examples that reflect an individual user's preferences. These examples can be extracted from the offline preference dataset for each user, providing concrete instances of preferred and dispreferred content. Conditioning on these examples allows for a more accurate estimation of the reward function $r(\rvc, \rvx_0, \rvu_i)$ for each user.

In this work, we utilize the features of a pre-trained vision-language Model (VLM), specifically LLaVA-OneVision~\citep{2024arXiv240803326L}, an open-source multimodal model with multi-image processing capabilities. For each user, a set of $N\!=\!4$ few-shot examples is selected, comprising a caption describing the context, a preferred image, and a dispreferred image. These examples are processed by the VLM, from which we extract an intermediate hidden state. 

To validate that the user embedding is effective, we condition on the frozen embedding to classify which user provided the preference annotations. As seen in Figure~\ref{fig:user-class}, the classifier significantly outperforms chance with a top-16 accuracy of 90\%, indicating that the embedding can disentangle different users in the preference dataset effectively. By incorporating this user-specific embedding into the conditioning of a text-to-image diffusion model, we enable user-aware reward modeling and generation.

\subsection{Personalization as Conditioning}
\label{sec:method-ipadapter}
In order to add the user embedding as conditioning to the diffusion model in image generation, we utilize additional cross-attention layers akin to IP-Adapter~\cite{ye2023ipadapter}. Here we learn to condition on the user information using the VLM embeddings and then integrate the image features into pretrained UNet by adapted modules with decoupled cross-attention. There are many other ways to add conditioning information to diffusion models; we chose this instantiation due to its training and parameter efficiency. 

We utilize Stable Cascade~\cite{pernias2024wrstchen} as our base model which comprises of three stages: Stages A-C. Stage A and B are image compressors, similar to the VAE in Stable Diffusion~\citep{rombach2022sd15} but achieve a much higher compression rate; and Stage C is the generation module conditioned on text. In Stage C, the text features $\rvc$ from pretrained CLIP text encoder are plugged into the model by feeding into the cross-attention layers. Given the query features $\rmZ$ and text features $\rvc_t$, the output is $\rmZ'\!=\!\text{Attention}(\rmQ, \rmK, \rmV)\!=\!\text{Softmax}(\frac{\rmQ\rmK^T}{\sqrt{d}})\rmV,
$
where $\rmQ\!=\!\rmZ\rmW_q$, $\rmK\!=\!\rvc_t\rmW_k$, $\rmV\!=\!\rvc_t\rmW_v$ are query, key and values, and $\rmW_q,\rmW_k,\rmW_v$ are trainable layers.

Following IP-Adapter~\citep{ye2023ipadapter}, we utilize a decoupled cross-attention mechanism to condition on the user embedding. First, we add a new cross-attention layer to each cross-attention layer in the Stage C model, using the original query $\rmQ$ and introducing new keys $\rmK'$ and values $\rmV'$ for user features. Then, we simply add the output of user cross-attention to the output of text cross-attention. Hence the final formulation of cross-attention is 

\begin{equation}
    \rmZ'=\text{Softmax}(\frac{\rmQ\rmK^T}{\sqrt{d}})\rmV + \text{Softmax}(\frac{\rmQ(\rmK')^T}{\sqrt{d}})\rmV',
\end{equation}

where $\rmQ\!=\!\rmZ\rmW_q$, $\rmK\!=\!\rvc_t\rmW_k$, $\rmV\!=\!\rvc_t\rmW_v$, $\rmK'\!=\!\rvu_t\rmW_k'$, $\rmV'\!=\!\rvu_t\rmW'_v$ and $\rmW_k',\rmW_v'$ are new introduced trainable parameters for $\rvu$.
During training, we optimize only the added cross-attention layers while keeping the parameters of pretrained diffusion model frozen, using the training objective as in \cref{eq:diff-dpo-u}. We also randomly drop user features (zero out the user embedding) in the training stage as regularization. This allows us to also prompt the model unconditionally of the user, by passing a zero embedding for the user.

\begin{table}[t]
\centering
\begin{tabular}{lccc}
\toprule
              & CLIP    & Aesthetic     & HPS  \\\midrule
Stable Cascade           &31.97          &5.33         &23.87        \\
\diffdpo   & 32.48		 &5.46         &25.96            \\
SFT   &32.26          &5.56         &25.78            \\
ours   &\textbf{32.66}          &\textbf{5.92}         &\textbf{27.51}               \\\midrule
\diffdpo (CLIP) &  32.96       &    -      &    -             \\
\diffdpo (Aesthetic)  &      -   &  6.42        &   -         \\
\diffdpo (HPS)  &       -  &       -   & 28.61              \\
\bottomrule
\end{tabular}

\caption{\textbf{Averaged scores on three reward function.} We condition on each reward to generate images both in SFT and our methods. \diffdpo (*) are upper-bounds.}
\vspace{-0.3cm}
\label{tab:synthetic}
\end{table}

\section{Experiments}
\label{sec:experiments}

\paragraph{Models and Dataset} In this section, we demonstrate the performance of \model across a range of experiments designed to assess its capabilities in personalized preference alignment. We use the objective from \cref{eq:diff-dpo-u} to fine-tune the open-source diffusion model, Stable Cascade~\cite{pernias2024wrstchen} -- specifically Stage C, which is responsible for text-conditioned generation. Following~\citet{wallace2024diffusion}, we fine-tune our model on the Pick-a-Pic v2 dataset~\cite{pickscore}, a preference dataset containing 58K text prompts and 0.8M image pairs labeled by 5K users.

\begin{figure}
    \centering
    \includegraphics[page=2,width=\linewidth]{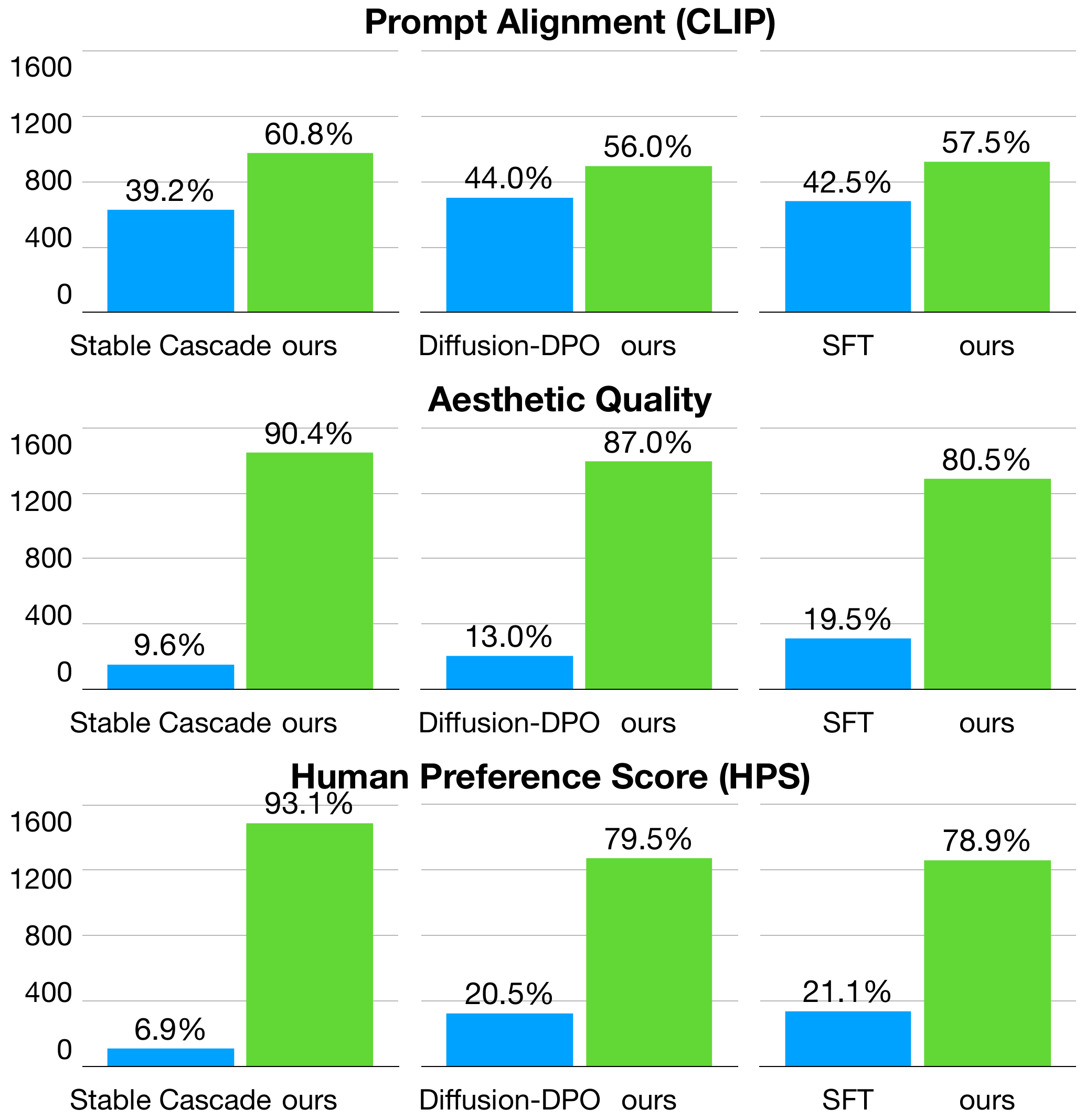}
    \caption{{\textbf{Automatic win rate evaluation with reward functions.} We compare against Stable Cascade, Diffusion-DPO, and SFT.}}
    \label{fig:exp1-win-rate}
\end{figure}

\begin{figure*}
    \centering
    \includegraphics[page=1,width=\linewidth]{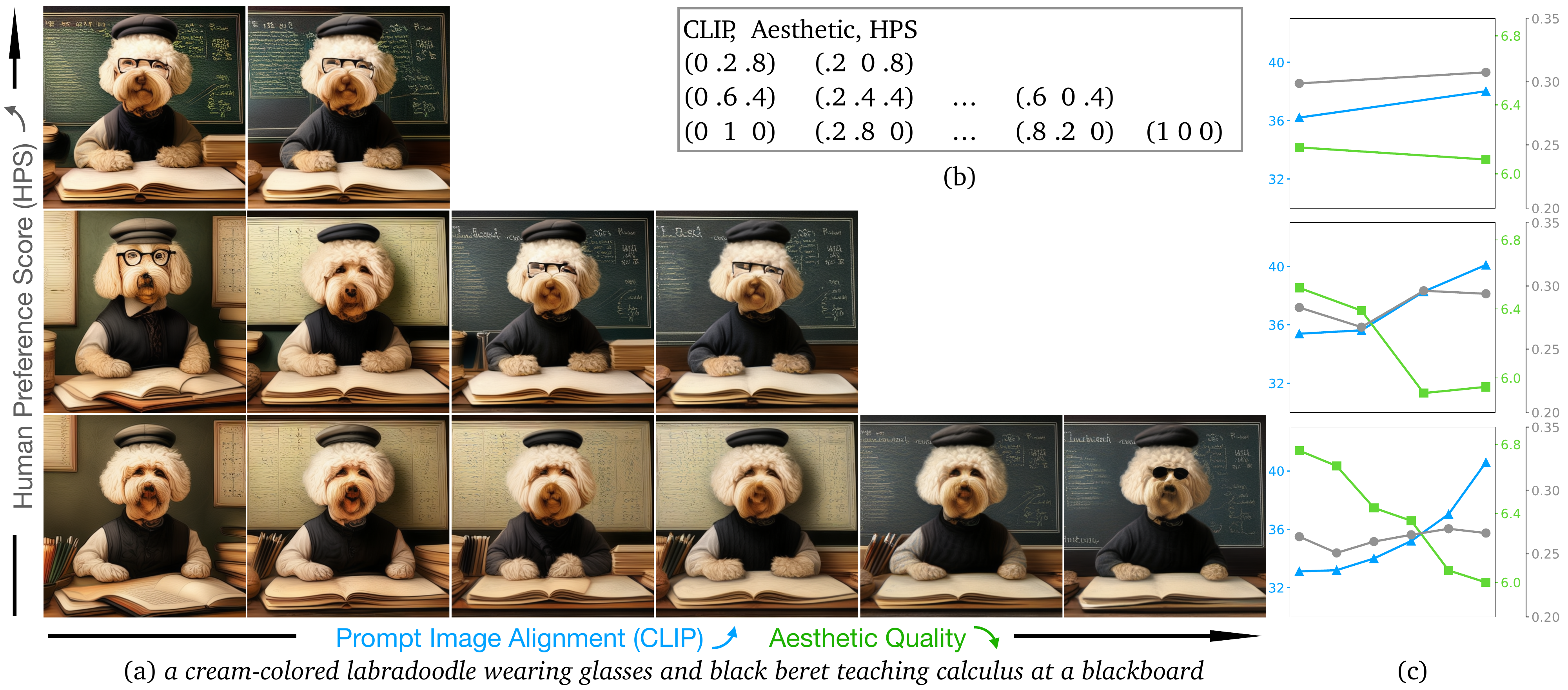}
    \caption{
\textbf{\model is able to interpolate among three distinct objectives during inference.} (a) generated images conditioned on (b) various weights, with three axes representing CLIP, Aesthetic, and HPS; (c) reward scores for each image. The score for each objective increases as its respective weight increases, and decreasing otherwise. For each row from left to right, the \textcolor{blue1}{CLIP} score increases alongside its weight, leading to a decrease in the \textcolor{green1}{Aesthetic} score as its weight decreases. From bottom to top, the \textcolor{grey1}{HPS} score increases with its weight.}
    \label{fig:synthetic}
\end{figure*}
\vspace{-0.1cm}
\paragraph{Training Details}  During training of \model, we only optimize the added cross-attention layers while keeping the pretrained diffusion model fixed, as described in \cref{sec:method-ipadapter}. We use AdamW~\cite{adamw} with an effective batch size of 768 pairs and a learning rate of $1 \times 10^{-5}$ for all experiments, training for one epoch. We find $\beta\in[0.1, 2]$ for hyperparameter tuning. Details are in Appendix~\ref{sec:app-exp}.
\vspace{-0.1cm}
\paragraph{User Information Conditioning} We begin by demonstrating the model’s ability to condition on synthetic user information, optimizing three distinct reward functions, we further show that generalize by linearly interpolating the weights of different reward functions (\cref{sec:exp1}). We then show that the model can also condition on real user preferences, generating personalized images based on few-shot examples from new users (\cref{sec:exp2}).

\begin{figure*}[t]
\centering
\setlength{\tabcolsep}{2pt} %
\begin{tabular}{ 0{p{0.20\textwidth}}  0{p{0.19\textwidth}}  0{p{0.19\textwidth}} 0{p{0.19\textwidth}} 0{p{0.19\textwidth}}}\toprule
\multicolumn{1}{c}{} & \multicolumn{1}{c}{\textbf{Stable Cascade}} & \multicolumn{1}{c}{\textbf{\diffdpo}} & \multicolumn{1}{c}{} & \multicolumn{1}{c}{} \\

\multicolumn{1}{c}{\multirow{-2}{*}{\textbf{User Profile}}} & \multicolumn{2}{c}{\cellcolor[HTML]{C0C0C0}{Append user profile to caption as augmentation}} & \multicolumn{1}{c}{\multirow{-2}{*}{\textbf{\diffdpo}}} &  \multicolumn{1}{c}{\multirow{-2}{*}{\textbf{Ours}}} \\

\hline

\footnotesize 
The preferred images share common themes of \textcolor{red}{wood grain, classic or futuristic design elements}, and a focus on aesthetics. They 
often feature unique front grille patterns, large windows, and a mix of traditional and modern design features. ...

&  \includegraphics[scale=0.18,valign=t]{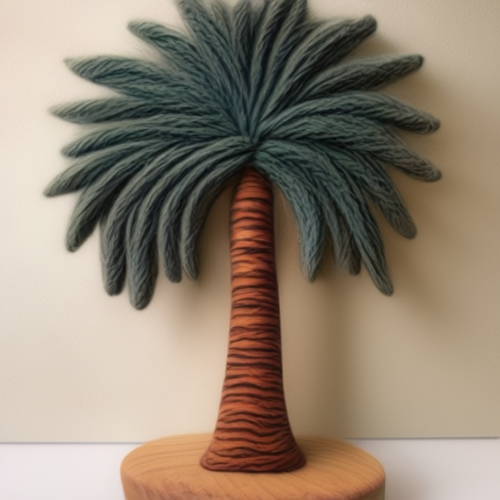} 
&  \includegraphics[scale=0.18,valign=t]{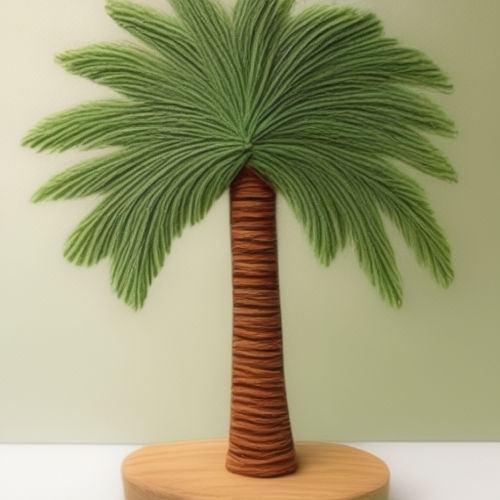}  
& \includegraphics[scale=0.18,valign=t]{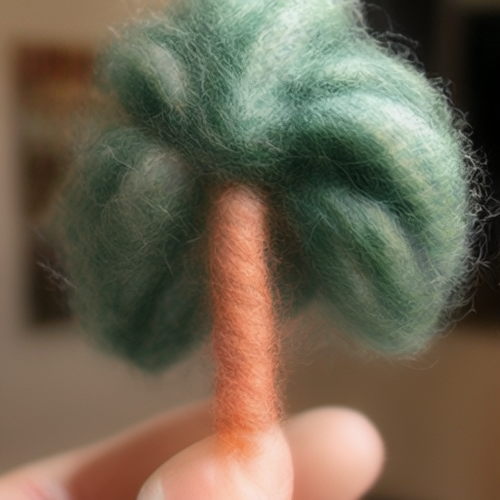} 
& \includegraphics[scale=0.18,valign=t]{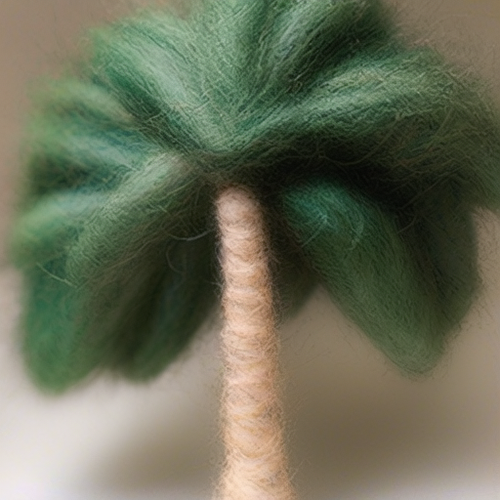} \\
\multicolumn{5}{c}{\emph{A needle-felted palm tree}}  \\ \hline

\footnotesize 
This user prefers images that have a personal and intimate feel \textcolor{ForestGreen}{with a focus on the subject's appearance and expression}. They also appreciate images with soft focus backgrounds and warm lighting that highlight texture and detail. ...

&  \includegraphics[scale=0.18,valign=t]{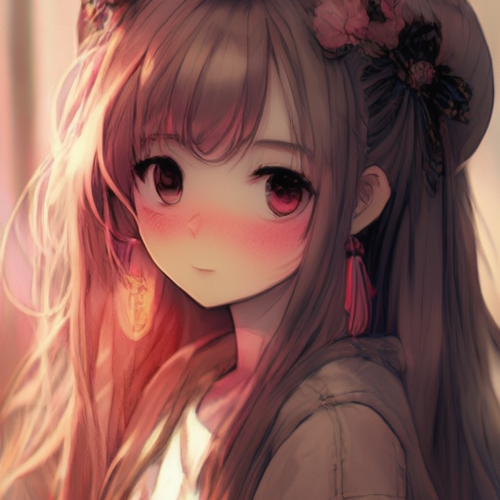} 
&  \includegraphics[scale=0.18,valign=t]{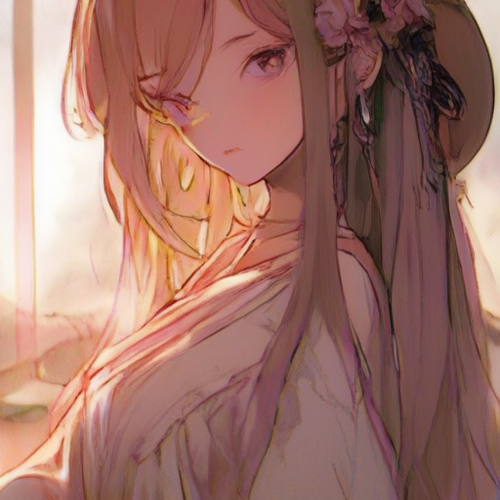}  
& \includegraphics[scale=0.18,valign=t]{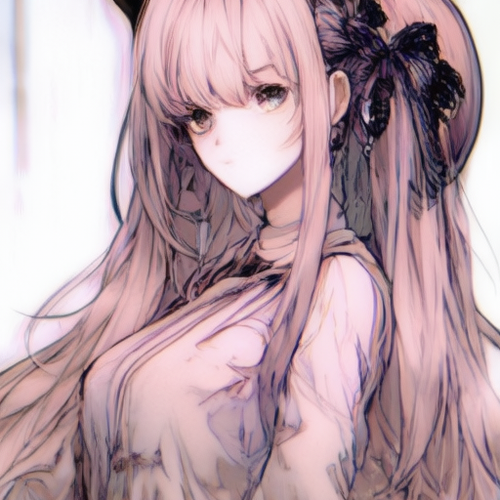} 
& \includegraphics[scale=0.18,valign=t]{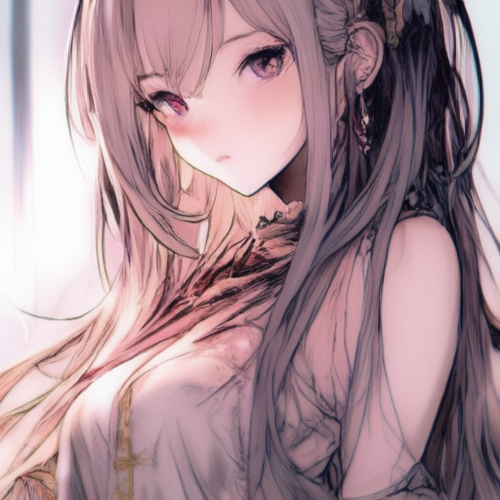} \\
\multicolumn{5}{c}{\emph{An anime girl, masterpiece, good line art, trending in pixiv}}  \\ \hline

\footnotesize The preferred images ...  such as a softer color palette, more dynamic lighting, and a \textcolor{ForestGreen}{balanced composition that enhances the visual appeal of the scene}. The characters are often positioned off-center, which adds to the dynamism of the composition. \textcolor{red}{The backgrounds are rich in detail.} ...

&  \includegraphics[scale=0.18,valign=t]{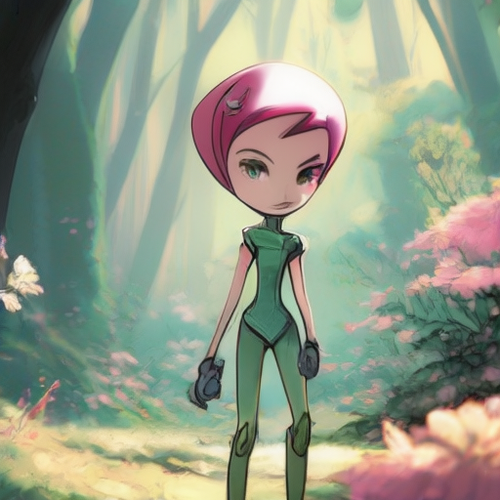} 
&  \includegraphics[scale=0.18,valign=t]{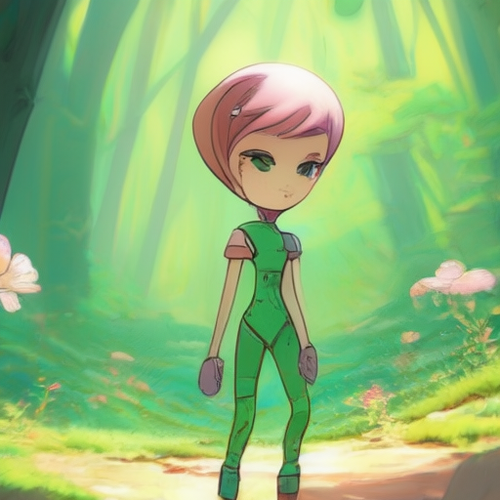}  
& \includegraphics[scale=0.18,valign=t]{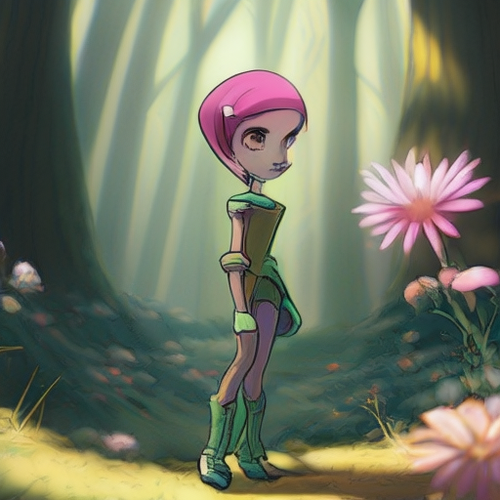} 
& \includegraphics[scale=0.18,valign=t]{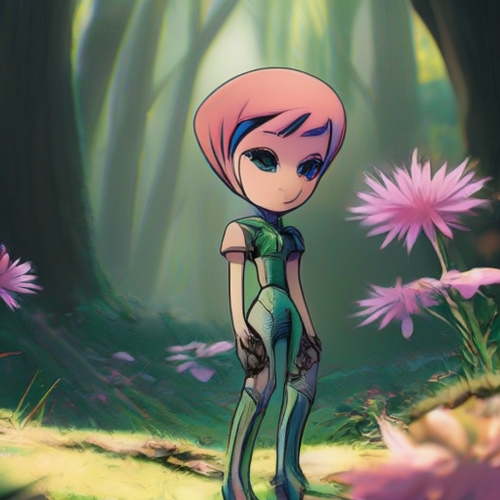} \\
\multicolumn{5}{c}{\emph{Aelita from Code Lyoko in the forest, gathering flowers}}  \\ \hline
\footnotesize 
The user is drawn to images that \textcolor{ForestGreen}{blend elements of reality with fantastical elements}, creating a sense of wonder and curiosity. ... The user may be interested in \textcolor{red}{fantasy art, animation, and storytelling through visual media.}

&  \includegraphics[scale=0.18,valign=t]{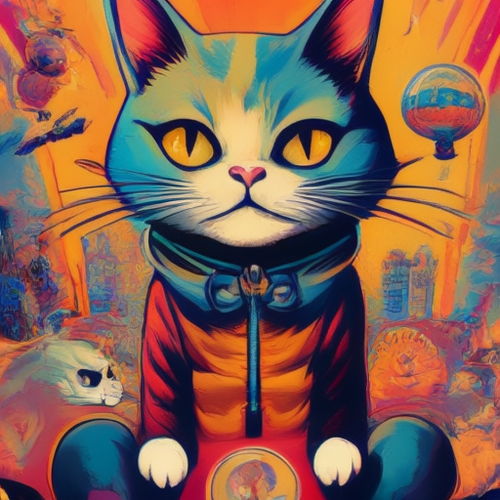} 
&  \includegraphics[scale=0.18,valign=t]{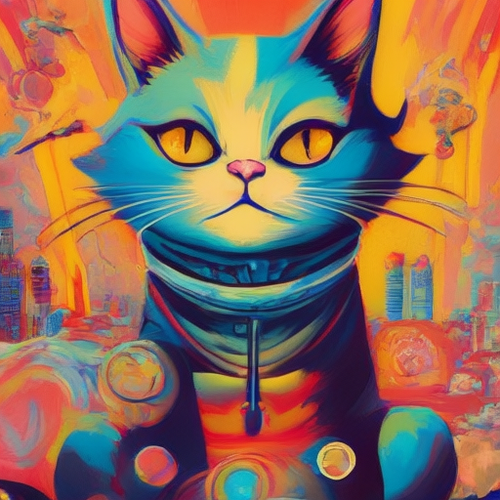}  
& \includegraphics[scale=0.18,valign=t]{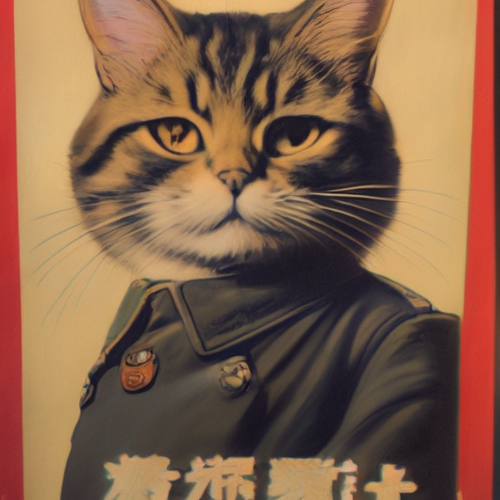} 
& \includegraphics[scale=0.18,valign=t]{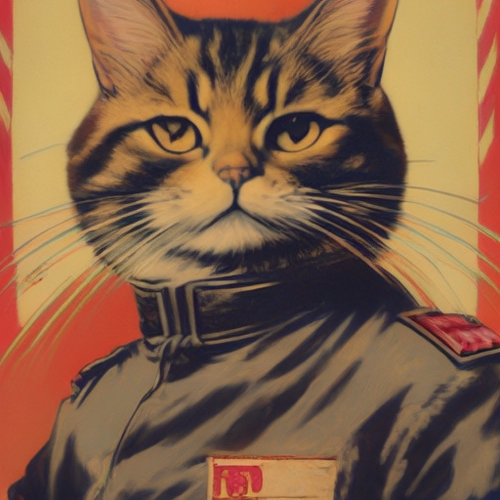} \\
\multicolumn{5}{c}{\emph{A cat on a propaganda poster}}  \\ 
\bottomrule
    \end{tabular}
    \caption{\textbf{Qualitative Analysis of Images Generated by \model and Baselines.} 
     Compared to \diffdpo, \model achieves closer alignment with the generated user profile, highlighted in \textcolor{ForestGreen}{green}. The caption-augmented method captures user profile details; however, it often leads to unintended image alterations, causing the image to disregard the caption itself, as indicated in \textcolor{red}{red}.}
    \label{fig:user_profile_qualitative}
\end{figure*}
\subsection{Synthetic User Alignment}
\label{sec:exp1}
\paragraph{Synthetic User} We utilize three reward functions, each representing a aspect of image-to-text generation quality, to simulate three distinct synthetic users: (i) CLIP (OpenCLIP ViT-H/14)~\cite{laionclip,clip} for prompt image alignment, (ii) Aesthetic Predictor~\cite{schuhmann22aes} for image visual appeal, and (iii) HPSv2 (Human Preference Score v2)~\cite{wu2023hpsv2} for human preference estimate. In the dataset, each user has an equal chance of appearing. We relabel ranking pairs $(\rvc, \rvxw_0, \rvxl_0, \rvu)$ based on the above reward functions, where $\rvu$ is a one-hot vector indicating the reward function, taking one of 3 values: $\rvu \in \{[1,0,0],[0,1,0], [0,0,1]\}$. 
\paragraph{Baseline} We compare our method with the following baseline: (1) Stable Cascade~\citep{pernias2024wrstchen} pretrained model, which is also the reference model during training; (2) \diffdpo, fine-tuned on Pick-a-Pic label without user conditioning~\cite{wallace2024diffusion}; (3) SFT, supervised fine-tuning on relabeled data with user conditioning. We also compare with (4) \diffdpo (CLIP/Aesthetic/HPS), each annotation indicating a separately fine-tuned model with corresponding reward, this serves as the upper-bound of our method, where oracle information about the user is provided during inference to choose the appropriate model. For a fair comparison, all baselines are fine-tuned using Stable Cascade with the same architecture as our model, optimizing only the added cross-attention layers. 

\paragraph{Text-to-Image Generation} For each method, we generate images with 30 inference steps using captions from PartiPrompts~\cite{yu2022partiprompt}, a benchmark containing 1.6K examples across various categories and challenge aspects.  For SFT and our method, we generate three images per prompt, each conditioned on one reward function.  For \diffdpo (CLIP/Aesthetic/HPS), images for each annotation are generated from its corresponding model. We report averaged reward scores in \cref{tab:synthetic} and win rate comparisons in~\cref{fig:exp1-win-rate}, we can see that our method significantly outperforms Stable Cascade in different reward functions, achieving win rates of 60\%, 90\%, and 93\% in terms of CLIP, Aesthetic, and HPS respectively. Our method effectively optimizes each reward by conditioning on individual preference, outperforming supervised fine-tuning by a large margin. Unlike \diffdpo, our method can jointly optimize multiple user rewards, achieving win rates comparable to \diffdpo models fine-tuned separately on each user's data.

\paragraph{Generalization via Interpolation}
During training we fine-tune the model using one-hot encoding of each objective. Specifically, the user embedding vector $\rvu$ being $[1,0,0],[0,1,0],[0,0,1]$ corresponding to optimization for CLIP, Aesthetic, and HPS. The user embedding can also be seen as  weighting mechanism for each objective. However, during inference, we are also interested in whether we can trade-off among different objectives by interpolating their respective weights. In \cref{fig:synthetic}, we generate images conditioned on various objective weightings, namely interpolating the given user embedding, we can see that as the weight of an objective increases, the corresponding score also rises, and vice versa. Therefore, our method can generalize to different reward combinations, even only trained on single objective data. This can not be achieved by optimizing separate models and each with an objective. 

This shows that by linearly interpolating the user embedding, we can adjust the rewards of the generated images accordingly. This finding aligns with the real user experiments presented next, as real user embeddings exhibit a linear correlation in embedding space.

\subsection{Personalized Real User Alignment}
\label{sec:exp2}
We now consider a realistic setting with preferences from the Pick-a-Pic dataset for each user. In this setting, we utilize the user features which are generated from a VLM on in-context preference examples as outlined in Section~\ref{sec:method-vlm}. We scale from the 3 synthetic "users" in Section~\ref{sec:exp1} to a filtered subset of the 4800 unique users in Pick-a-Pic. 

\paragraph{Experiment setting}
We repartition the Pick-a-Pic dataset into a set of 4 splits: (1) Train, (2) Validation, (3) Test with Seen Users + Held-out Captions and (4) Test with Unseen Users + Held-out Captions. This decomposition allows us to study independently the generalization of our methods to new prompts and unseen users. Additionally, for each user, we take sets of $N\!=\!4$ few-shot preference examples for each user and use these groupings for the conditioning of the VLM on few-shot preference pairs. We thus have to filter out 824 users in Pick-a-Pic which have less than 4 examples per user, as no user embedding for this user can be formed. We generate the user embedding through the procedure outlined in~\cref{sec:method-vlm}. 

\paragraph{Generation Results} We perform qualitative analysis on the set of images generated for each user. We perform this evaluation by generating a user profile per user. We use a set of few-shot preference pairs per user which we condition a VLM on along with Chain-of-Thought prompting (COT) to construct a user-profile consistent with the preference pairs. The COT prompt asks the model to summarize each image in the preference pair and describe the differences between them, which is then aggregated to form a user profile. Sample user profiles and the full Chain-of-Thought prompt is found in Appendix~\ref{sec:app-exp}. As seen in ~\cref{fig:user_profile_qualitative}, our method has closer alignment to the user profile with respect to Stable Cascade and Diffusion-DPO, being able to pick up details about the user such as the highlighted part. While Stable Cascade and \diffdpo with augmented captions can capture certain aspects of the user description, they often overfit to the user profile, resulting in generated images that overlook the intended image caption. 

\paragraph{Personalized Preference Alignment Evaluation} We evaluate whether the generated model aligns with specific user preference using a VLM as a judge, leveraging strong VLMs such as GPT4o-vision~\citep{OpenAI2023GPT4TR}, similar to recipes in prior work~\cite{2024arXiv240106591L, 2024arXiv240704842C}. In particular, we condition the VLM on few-shot examples from the user as well as a pair of images for a given test time caption and ask the model to judge which of the two images would be preferred by the user. To ensure consistency of the response the model selects, we judge each pair of images twice using the VLM, flipping the ordering of the images, filtering out pairs where the VLM is inconsistent. With this paradigm and $4$ shots per user, we find that the VLM is able to match the ground truth personalized human preference with $83\%$ accuracy.

We decompose our evaluation into two settings: (1) evaluating users that were seen during training but on captions that were held out and (2) evaluating held-out users and captions. We aggregate over these settings to also compute metrics for the general user in the dataset. For the 4341 users in Pick-a-Pic we select 192 held out and  277 in-distribution users. For each of these users, we choose $4+1$ examples, where 4 examples are used for the in-context conditioning for the user and 1 example is used as the caption for evaluation. We compute an automatic win-rate using pairs of models, generating one image for each caption and evaluating with the VLM-as-Judge pipeline outlined above.

\begin{figure}
    \centering
    \includegraphics[width=\linewidth]{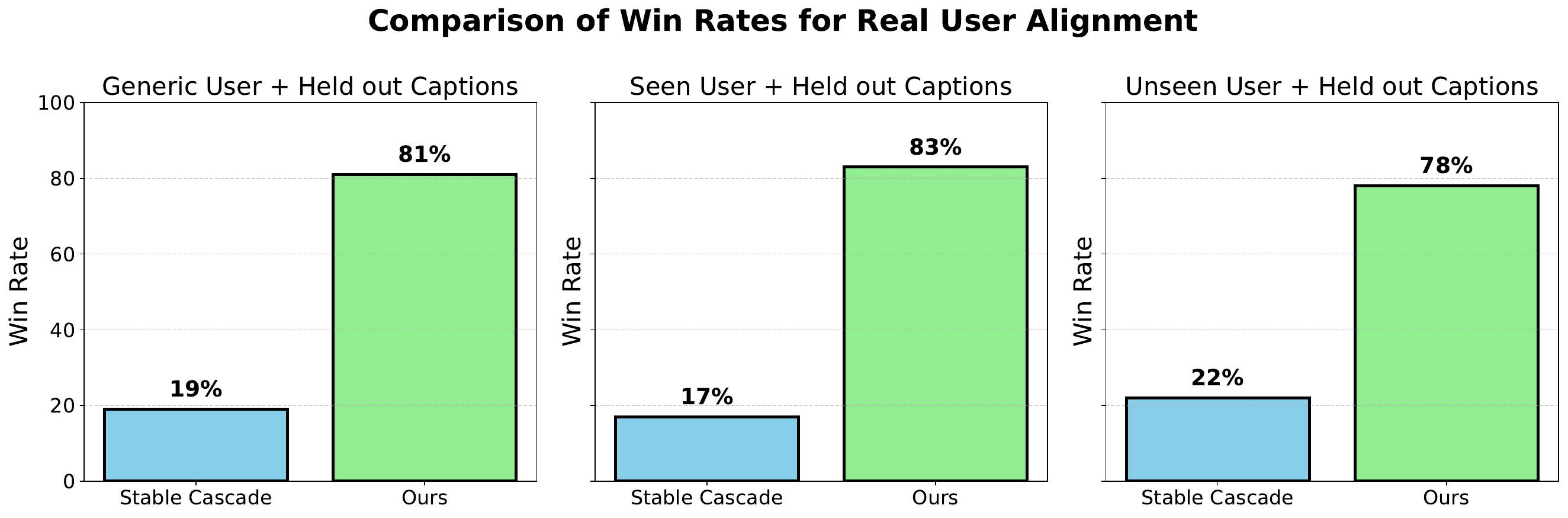}
    \caption{\textbf{Automatic Win Rate Evaluation with GPT-4o} We evaluate on seen users, held out users, and aggregate to see win-rates for generic users, comparing Stable Cascade and \model.}
    \label{fig:win-rate-4o}
\end{figure}

We report win rate comparisons in ~\cref{fig:win-rate-4o}. Here, \model significantly outperforms Stable Cascade in all three evaluation settings, achieving a win rate of 83\% on seen users, and 78\% on unseen users, aggregated to 81\% on all users. 

\section{Discussion, Conclusion, and Limitations}
\label{sec:discussion}

This work presents a framework for personalization in text-to-image diffusion models using user feature embeddings derived from a VLM. By conditioning on these embeddings through decoupled cross-attention layers and fine-tuning with the DPO objective, our method achieves both accurate reward modeling and high-quality personalized image generation. This approach allows us to generalize effectively to unseen users, requiring only few-shot preference examples from new users at test time.

However, several open questions and limitations remain. Currently, our conditioning on few-shot examples is caption agnostic, as it selects a set of random preferences from a user. Future work could investigate methods for actively selecting the preferences that the VLM embeds, potentially enhancing personalization. Additionally, extending this technique to video generation could enable effective personalized video content, which is an exciting direction for further research. Finally, exploring the trade-offs between implicit and stated preferences may provide insights into optimizing user experience and model performance.

\section*{Acknowledgment}
This research is done during MD's internship at Luma AI and supported in part by ARO (W911NF-21-1-0125), ONR (N00014-23-1-2159), and the CZ Biohub. AS gratefully acknowledges the support of the NSF Graduate Research Fellowship Program.

{
    \small
    \bibliographystyle{ieeenat_fullname}
    \bibliography{main}
}
\clearpage
\setcounter{page}{1}

\appendix
\maketitlesupplementaryonecolumn

\section{Direct Preference Optimization on Diffusion Models}
\label{sec:app-diffusion}
Given dataset containing examples $(\rvc, \rvxw_0, \rvxl_0)$, we define $r(\rvc,\rvx_0)$ as the reward on image $\rvx_0$ given prompt $\rvc$. We would like to fine-tune a text-to-image models $p_{\theta}(\rvx_{0}|\rvc)$ such that reward is maximized while keeping close to a reference model $p_{\text{ref}}(\rvx_0|\rvc)$ in terms of KL-divergence as regularization :
\begin{equation}
    \max_{p_{\theta}} \E_{\rvc, \rvx_0} \left[r(\rvc,\rvx_0)\right] -\beta \KL \left[p_{\theta}(\rvx_{0}|\rvc)\|p_{\text{ref}}(\rvx_0|\rvc)\right],
    \label{eq:obj-dpo}
\end{equation}
where $\beta$ is a parameter controlling how much $p_{\theta}(\rvx_{0}|\rvc)$ deviates from $p_{\text{ref}}(\rvx_0|\rvc)$.

We introduce latent variables $\rvx_{1:T}$ and define $R(\rvc,\rvx_{0:T})$ as the reward on the whole diffusion chain, such that we can define $r(\rvc, \rvx_0) = \E_{\p(\rvx_{1:T}|\rvx_0, \rvc)}[R(\rvc, \rvx_{0:T})]$.  Given \cref{eq:obj-dpo}, we have 
\begin{align}
\begin{split}
    \min_{p_\theta} & - \E_{p_\theta(\rvx_0|\rvc)}\left[r(\rvc, \rvx_0)/\beta\right] + 
    \KL
    \left[p_\theta(\rvx_{0}|\rvc)||p_{\text{ref}}(\rvx_{0}|\rvc)\right]\\
    \leq \min_{p_\theta} & - \E_{p_\theta(\rvx_0|\rvc)}[r(\rvc, \rvx_0)/\beta] + 
    \KL\left[p_\theta(\rvx_{0:T}|\rvc)||p_{\text{ref}}(\rvx_{0:T}|\rvc)\right] \\
    = \min_{p_\theta} & - \E_{p_\theta(\rvx_{0:T}|\rvc)}[R(\rvc,\rvx_{0:T})/\beta] + 
    \KL\left[p_\theta(\rvx_{0:T}|\rvc)||p_{\text{ref}}(\rvx_{0:T}|\rvc)\right] \\
    = \min_{p_\theta} & \ \E_{p_\theta(\rvx_{0:T}|\rvc)}\left(\log\frac{p_\theta(\rvx_{0:T}|\rvc)}{p_{\text{ref}}(\rvx_{0:T}|\rvc)\exp(R(\rvc,\rvx_{0:T})/\beta)/Z(\rvc)} - \log Z(\rvc)\right) \\
    = \min_{p_\theta}  & \ \KL\left[p_\theta(\rvx_{0:T}|\rvc)\|p_{\text{ref}}(\rvx_{0:T}|\rvc)\exp(R(\rvc, \rvx_{0:T})/\beta)/Z(\rvc)\right].
    \label{eq:min_joint_kl}
\end{split}
\end{align}
where $Z(\rvc)=\sum_{\rvx}p_{\text{ref}}(\rvx_{0:T}|\rvc)\exp\left(r(\rvc,\rvx_0)/\beta\right)$ is the partition function.
The optimal $p_\theta^*(\rvx_{0:T}|\rvc)$ of Equation~(\ref{eq:min_joint_kl}) has a unique closed-form solution:
\begin{align*}
    p_\theta^*(\rvx_{0:T}|\rvc) = p_{\text{ref}}(\rvx_{0:T}|\rvc) \exp(R(\rvc,\rvx_{0:T})/\beta) / Z(\rvc),
\end{align*}
Therefore, we have the reparameterization of reward function
\begin{align*}
    R(\rvc, \rvx_{0:T}) & = \beta \log\frac{p_\theta^*(\rvx_{0:T}|\rvc)}{p_{\text{ref}}(\rvx_{0:T}|\rvc)} + \beta \log Z(\rvc).
\end{align*}
Plug this into the definition of $r$, hence we have 
\begin{equation*}
r(\rvc,\rvx_0)=\beta \E_{p_\theta(\rvx_{1:T}|\rvx_0,\rvc)}\left[ \log \frac{p^*_{\theta}(\rvx_{0:T}|\rvc)}{p_{\text{ref}}(\rvx_{0:T}|\rvc)}\right] + \beta\log Z(\rvc).
\end{equation*}
Substituting this reward reparameterization into maximum likelihood objective of the Bradly-Terry model, the partition function cancels for image pairs, and we get a maximum likelihood objective defined on diffusion models, for a single pair $(\rvc, \rvxw_0, \rvxl_0)$: 
\begin{equation*}
    L_\text{\diffdpo}(\theta) =
    -\log \sigma \left(\beta
    {\mathbb{E}}_{{\rvxw_{1:T},\rvxl_{1:T}}}\left[ \log\! \frac{p_{\theta}(\rvxw_{0:T}|\rvc)}{p_{\text{ref}}(\rvxw_{0:T}|\rvc)}- \log\! \frac{p_{\theta}(\rvxl_{0:T}|\rvc)}{p_{\text{ref}}(\rvxl_{0:T}|\rvc)}\right]\right)%
\end{equation*}
where $\rvxw_{1:T}\sim p_\theta(\rvx_{1:T}|\rvxw_0,\rvc)$ and $\rvxl_{1:T} \sim p_\theta(\rvx_{1:T}|\rvxl_0,\rvc) $.
Since sampling from $p_\theta(\rvx_{1:T}|\rvx_0, \rvc)$ is intractable, we utilize the forward process $q(\rvx_{1:T}|\rvx_0)$ for approximation. 
\begin{align}
\begin{split}\label{eq-app:l1}
L_\text{approx}(\theta) & =
    -\log \sigma \left(\beta
    {\mathbb{E}}_{{\rvxw_{1:T},\rvxl_{1:T}  }}\left[ \log\! \frac{p_{\theta}(\rvxw_{0:T})}{p_{\text{ref}}(\rvxw_{0:T})}- \log\! \frac{p_{\theta}(\rvxl_{0:T})}{p_{\text{ref}}(\rvxl_{0:T})}\right]\right) \\
\end{split}
\end{align}
where $ \rvxw_{1:T}\sim q(\rvx_{1:T}|\rvxw_0,\rvc), \rvxl_{1:T} \sim q(\rvx_{1:T}|\rvxl_0,\rvc)$. Therefore,
\begin{align}
\begin{split}
L_\text{approx}(\theta) 
 & = -\log \sigma \left(\beta
    {\mathbb{E}}_{{\rvxw_{1:T},\rvxl_{1:T}  }}\left[ \sum_{t=1}^T \log\! \frac{p_{\theta}(\rvxw_{t-1}|\rvxw_t)}{p_{\text{ref}}(\rvxw_{t-1}|\rvxw_t)}- \log\! \frac{p_{\theta}(\rvxl_{t-1}|\rvxl_t)}{p_{\text{ref}}(\rvxl_{t-1}|\rvx_t)}\right]\right)\\
 & = -\log \sigma \left(\beta
    {\mathbb{E}}_{{\rvxw_{1:T},\rvxl_{1:T}  }} T \E_t \left[ \log \frac{p_{\theta}(\rvxw_{t-1}|\rvxw_t)}{p_{\text{ref}}(\rvxw_{t-1}|\rvxw_t)}- \log \frac{p_{\theta}(\rvxl_{t-1}|\rvxl_t)}{p_{\text{ref}}(\rvxl_{t-1}|\rvx_t)}\right]\right) \\
 & = -\log \sigma \left(\beta T 
     \E_{t, \rvxw_{t-1,t}, \rvxl_{t-1,t}}\left[ \log \frac{p_{\theta}(\rvxw_{t-1}|\rvxw_t)}{p_{\text{ref}}(\rvxw_{t-1}|\rvxw_t)}- \log \frac{p_{\theta}(\rvxl_{t-1}|\rvxl_t)}{p_{\text{ref}}(\rvxl_{t-1}|\rvx_t)}\right]\right) \\
 & = -\log \sigma \Biggl(\beta T 
     \E_{{t, \rvxw_{t},\rvxl_{t},  \rvxw_{t-1},\rvxl_{t-1}}}\left[ \log \frac{p_{\theta}(\rvxw_{t-1}|\rvxw_t)}{p_{\text{ref}}(\rvxw_{t-1}|\rvxw_t)}- \log \frac{p_{\theta}(\rvxl_{t-1}|\rvxl_t)}{p_{\text{ref}}(\rvxl_{t-1}|\rvx_t)}\right]\Biggr)
\end{split}
\end{align}
where $\rvxw_{t} \sim q(\rvx_{t}|\rvxw_0), \rvxl_{t} \sim q(\rvx_{t}|\rvxl_0)$ and $ \rvxw_{t-1}\sim q(\rvx_{t-1}|\rvxw_t,\rvxw_0),\rvxl_{t-1}\sim q(\rvx_{t-1}|\rvxl_t,\rvxl_0)$. Since function $-\log \sigma$ is a convex function, by Jensen's inequality, we can push $\E_{{t, \rvxw_{t},\rvxl_{t}}}$ to the outside of $-\log \sigma$ and get an upper bound, therefore we have
\begin{align*}
    L_\text{approx}(\theta)  \leq & 
- \E_{{t,  \rvxw_{t},\rvxl_{t}}} \log \sigma \left(
\beta T 
        \E_{\rvxw_{t-1},\rvxl_{t-1}}\left[ \log \frac{p_{\theta}(\rvxw_{t-1}|\rvxw_t)}{p_{\text{ref}}(\rvxw_{t-1}|\rvxw_t)}- \log \frac{p_{\theta}(\rvxl_{t-1}|\rvxl_t)}{p_{\text{ref}}(\rvxl_{t-1}|\rvx_t)}\right]\right) \\
=&  - \E_{{t,\rvxw_{t},\rvxl_{t}}} \log\sigma \Biggl( -\beta T \biggl(
 \left(\KL[q(\rvxw_{t-1}|\rvxw_{0,t})\|p_\theta(\rvxw_{t-1}|\rvxw_t)] 
 - \KL[q(\rvxw_{t-1}|\rvxw_{0,t})\|p_{\text{ref}}(\rvxw_{t-1}|\rvxw_t)] \right) \\
   & \hspace{3.4cm}- \left( \KL[q(\rvxl_{t-1}|\rvxl_{0,t})\|p_\theta(\rvxl_{t-1}|\rvxl_t)]
 - \KL[q(\rvxl_{t-1}|\rvxl_{0,t})\|p_{\text{ref}}(\rvxl_{t-1}|\rvxl_t)]\right) \biggr) \Biggl)
 \end{align*}

Using the Gaussian parameterization of the reverse process (\cref{eq:dm}), the above loss simplifies to:
\begin{align*}
    L_\text{approx}(\theta)
=  - \E_{\rvc,\rvxw_0, \rvxl_0} \log\sigma \Biggl( -\beta T \omega(\lambda_t)  & \biggl(
\| \rvepsilonw -\rvepsilon_\theta(\rvxw_t,\rvc,t)\|^2_2 - \|\rvepsilonw - \rvepsilon_\text{ref}(\rvxw_t,\rvc,t)\|^2_2 \\
   & - \left( \| \rvepsilonl \!-\rvepsilon_\theta(\rvxl_{t}\!,\rvc,t)\|^2_2 - \|\rvepsilonl\! - \rvepsilon_\text{ref}(\rvxl_{t}\!,\rvc,t)\|^2_2\right) \biggr) \Biggl)
 \end{align*}
where $\rvepsilonw, \rvepsilonl \sim \mathcal{N}(\mathbf{0}, \mathbf{I})$, $\rvxw_{t} \sim q(\rvxw_t|\rvx_0)$,  $\rvxl_{t} \sim q(\rvxl_t|\rvx_0)$, $\lambda_t=\alpha_t^2/\sigma_t^2$ is a signal-to-noise ratio term.

\newpage

\section{Experiments Details}
\label{sec:app-exp}
\subsection{Diffusion Model Fine-tuning Details}
For the experiments shown in \cref{sec:exp1}, we optimize only the added cross-attention layers,  with 150M trainable parameters in total. All models are trained using the AdamW optimizer with an effective batch size of 768 pairs, a learning rate of $1 \times 10^{-5}$, and a single training epoch. The hyperparameter $\beta$ is tuned within the range $[0.1, 2]$. Training is conducted on the Pick-a-Pic training set, with the best $\beta$ selected based on the averaged rewards of generated samples evaluated on the Pick-a-Pic validation set, which contains 500 unique captions. Results are reported on the Partipromt dataset containing 1632 captions. For each method (each row in \cref{tab:synthetic}), $\beta$ is tuned independently. Experiments are conducted on H100 GPUs with 80GB of memory. Using 8 GPUs, each with a local batch size of 16, training for one epoch (approximately 1000 gradient update steps) takes about 2 hours.

For the personalized real user experiments shown in \cref{sec:exp2}, we use the same training settings as in \cref{sec:exp1} except for a reduced learning rate of $3 \times 10^{-6}$. We train on Pick-a-Pic training set, find the best hyper-parameter on Pick-a-Pic validation set and report results on Partiprompt dataset. As there are no automatic evaluation metrics to evaluate alignment with individualized user preferences, we find the best $\beta$ based on the highest PickScore on the Pick-a-Pic validation set. We then run \diffdpo with the same $\beta$ as baseline. 
\subsection{Additional Details for Generating User Embeddings}
As mentioned in \cref{sec:method-vlm}, features from pre-trained VLM, LLaVA-OneVision~\citep{li2024llava}, are constructed from $N=4$ few-shot examples. To elicit these features, we employ Zero-Shot Chain of Thought Prompting (COT)~\citep{kojima2023largelanguagemodelszeroshot,wei2023chainofthoughtpromptingelicitsreasoning} to allow the model to reason about the images in a preference pair as well as generate a user profile. The prompt used for this COT can be found in \cref{tab:cot}. We then extract an embedding from the VLM from the last hidden state of the Qwen 2 Language Model in the LLaVA OneVision architecture for the final token it generates for the User Profile (Step 5 in the COT Assistant Prompt in \cref{tab:cot}). For sampling, we use a temperature of 0.7 with nucleus sampling probability of 1.0 (no nucleus sampling). As seen in \cref{fig:user-class}, the top-k accuracy of a learned classifier from this frozen embedding is high, significantly outperforming random chance, indicating that this embedding is expressive, able to distinguish users within the pick-a-pick dataset. We additionally store the generated user profile for the baselines where the user profile is appended to the caption as an augmentation for the fairest comparison.

\subsection{Additional Details for Scoring}
We similarly employ COT for scoring. We add an additional assistant prompt for User Preference Prediction as found in \cref{tab:cot}. Here, we employ a stronger VLM as a Judge, GPT 4o-mini. For consistency, we present each pair of images twice to the model. In particular, for two images A and B, we ask the VLM to judge the images in two different permutations: first A then B and first B then A. We omit comparisons where the model isn't consistent in scoring (i.e A chosen for both comparisons or B chosen for both comparisons). For sampling, we utilize a temperature of 1.0 and nucleus sampling probability of 1.0 (no nucleus sampling). With COT and consistency, we find that we can match the preferences from real users in the Pick-a-Pic v2 dataset~\citep{pickscore} with 83\% accuracy. 

\subsection{Additional Details for Evaluation and Dataset Construction} Due to the restriction of fewshot prompting, we require at least $N=4$ examples per user. Therefore, we drop users in the Pick-a-Pic v2 dataset where the number of preference pairs that the user labels are below $N$. For classification, we subsample 300 users with the most preference pairs to allow for more examples in classification.

\newpage

\begin{table}[h!]
\centering
\begin{tabular}{|p{15cm}|}
\hline
\textbf{System Prompt} \\ \hline
You are an expert in image aesthetics and have been asked to predict which image a user would prefer based on the examples provided. \\ \hline
\textbf{COT Assistant Prompt} \\ \hline
You will be shown a few examples of preferred and dispreferred images that a user has labeled. \\
\cr
Here is Pair 1: \\
Here is the caption: [Caption for Pair 1] \\
Here is Image 1: [Image 1] \\
Here is Image 2: [Image 2] \\
Prediction of user preference: [1 or 2] \\
\cr
[...] \\
\cr
Here is Pair 4: \\
Here is the caption: [Caption for Pair 1] \\
Here is Image 1: [Image 1] \\
Here is Image 2: [Image 2] \\
Prediction of user preference: [1 or 2] \\
\cr
1. Describe each image in terms of style, visual quality, and image aesthetics. \\ 
2. Explain the differences between the two images in terms of style, visual quality, and image aesthetics. \\ 
3. After you have described all of the images, summarize the differences between the preferred and dispreferred images into a user profile. \\ 
\cr
Format your response as follows for the four pairs of images: \\ 
\cr
Pair 1: \\
Image 1: [Description] \\
Image 2: [Description] \\
Differences: [Description] \\
\cr
[...] \\
\cr
Pair 4: \\
Image 1: [Description] \\
Image 2: [Description] \\
Differences: [Description] \\
\cr
User Profile: [Description] \\ \hline

\textbf{Additional Assistant Prompt for User Preference Prediction} \\ \hline

Finally, you are provided with a new pair of images, unlabeled by the user. Your task is to predict which image the user would prefer based on the previous examples you have seen. \\ 
Format your response as follows: \\ 
Prediction of user preference: [1 or 2] \\ \hline
\end{tabular}
\caption{Instructions for Embedding Generation and User Preference Prediction}
\label{tab:cot}
\end{table}

\clearpage
\section{More Qualitative Examples}
\label{sec:app-examples}

Similar to \cref{fig:synthetic}, the following figures show that \model is able to interpolate among three distinct rewards during inference. 
\begin{figure*}[ht]
\centering
\begingroup
\setlength{\tabcolsep}{0pt} %
\renewcommand{\arraystretch}{-100} %
\begin{tabular}{ 0{p{0.166\textwidth}}  0{p{0.166\textwidth}}  0{p{0.166\textwidth}} 0{p{0.166\textwidth}} 0{p{0.166\textwidth}} 0{p{0.166\textwidth}} }\toprule

\includegraphics[scale=0.161,valign=t]{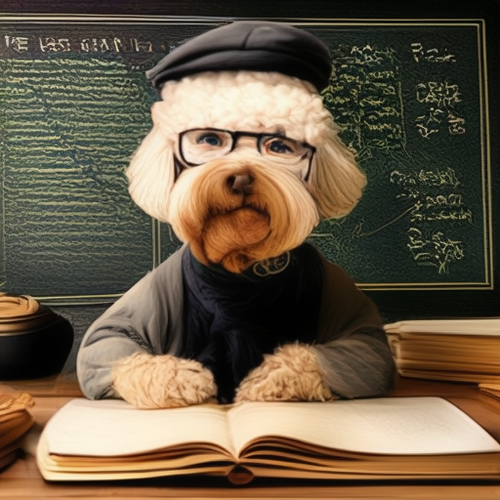}\\
\includegraphics[scale=0.161,valign=t]{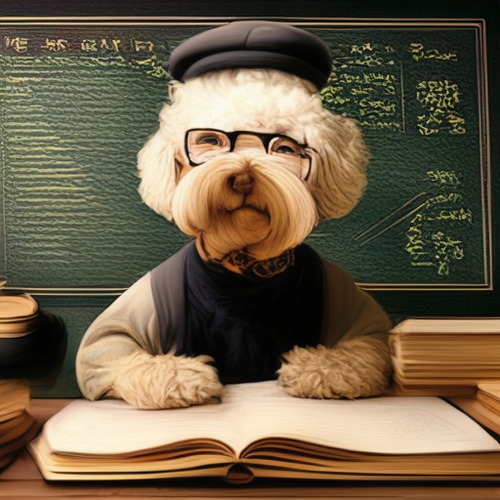} &
\includegraphics[scale=0.161,valign=t]{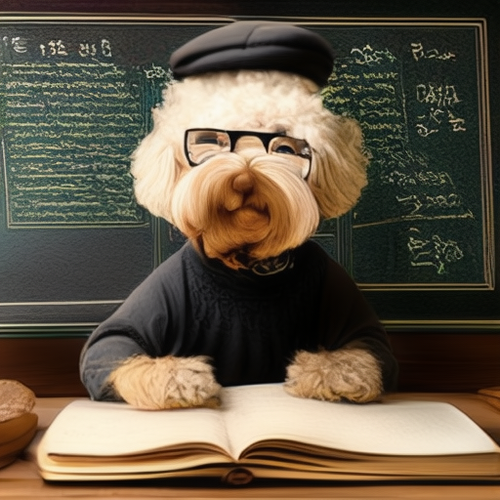}\\
\includegraphics[scale=0.161,valign=t]{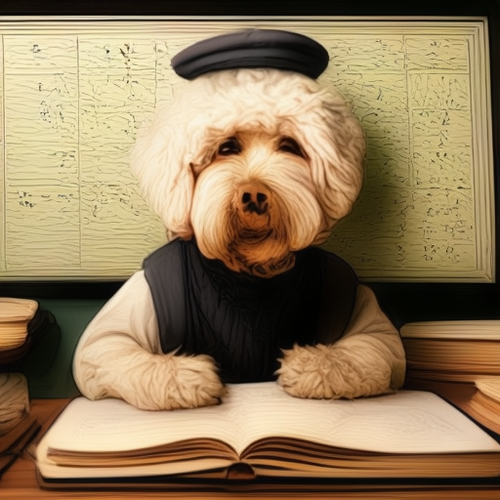} &
\includegraphics[scale=0.161,valign=t]{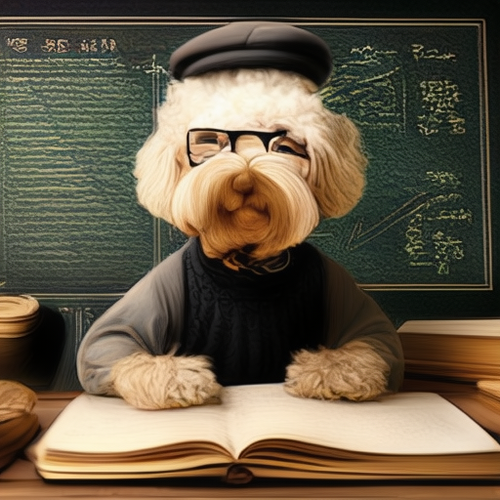} &
\includegraphics[scale=0.161,valign=t]{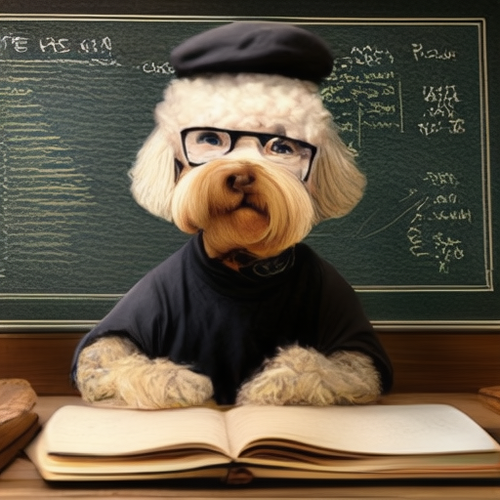}\\
\includegraphics[scale=0.161,valign=t]{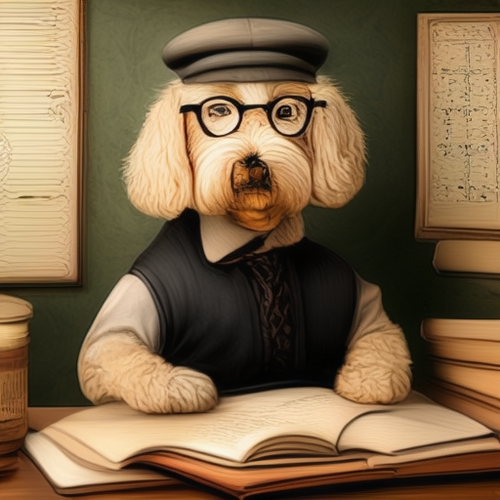} &
\includegraphics[scale=0.161,valign=t]{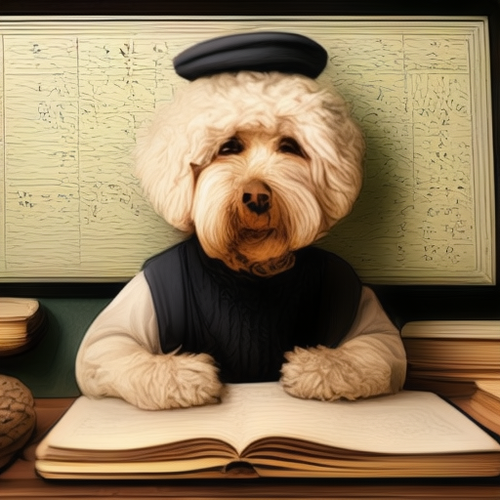} &
\includegraphics[scale=0.161,valign=t]{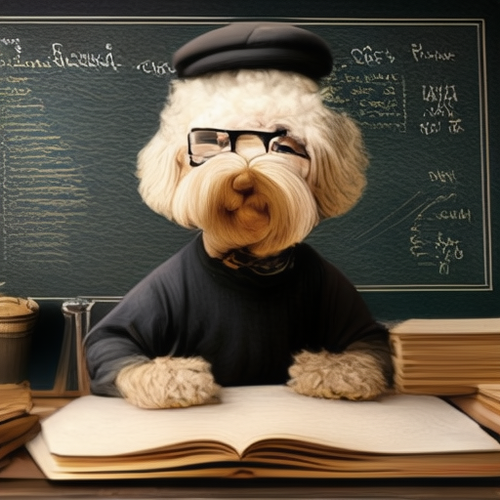} &
\includegraphics[scale=0.161,valign=t]{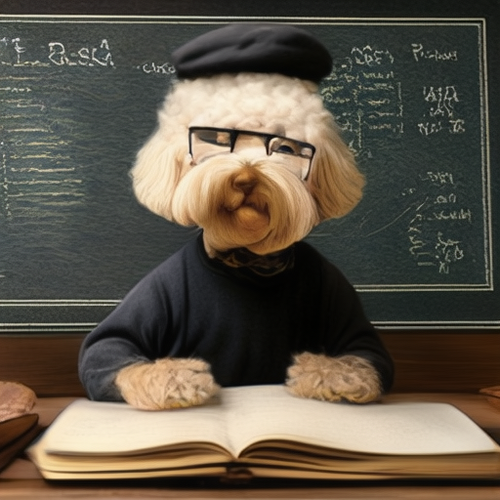} \\
\includegraphics[scale=0.161,valign=t]{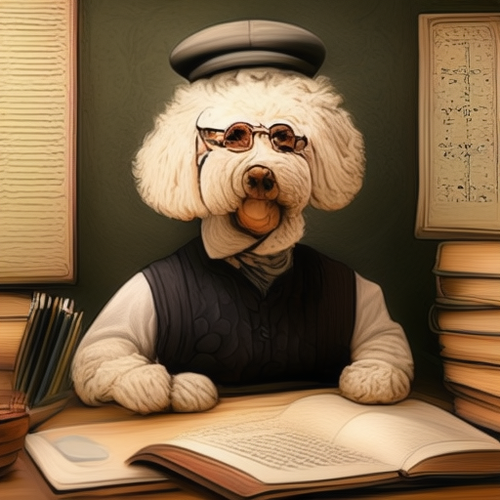} &
\includegraphics[scale=0.161,valign=t]{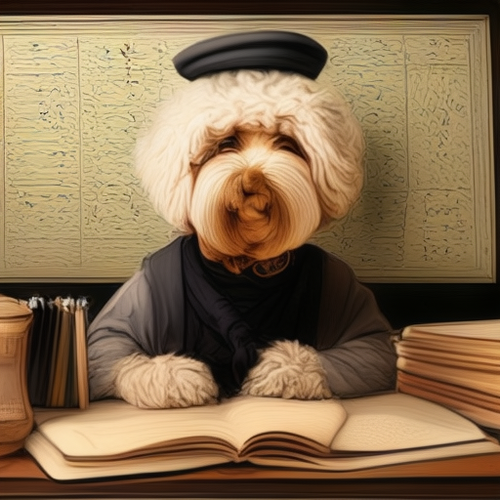} &
\includegraphics[scale=0.161,valign=t]{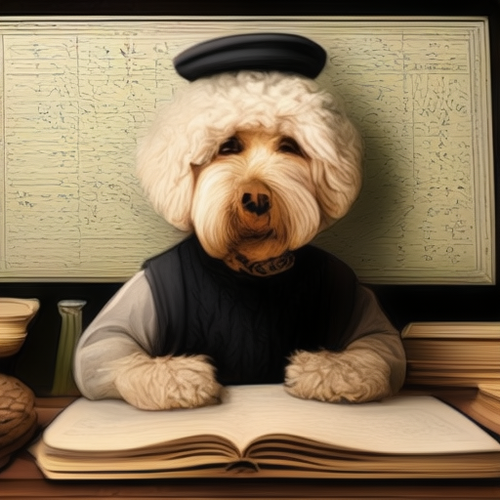} &
\includegraphics[scale=0.161,valign=t]{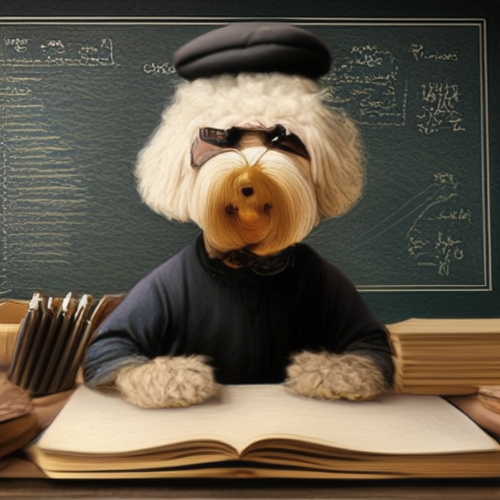} &
\includegraphics[scale=0.161,valign=t]{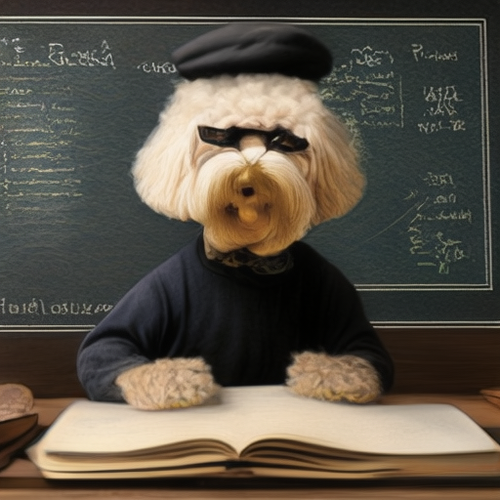} \\
\includegraphics[scale=0.161,valign=t]{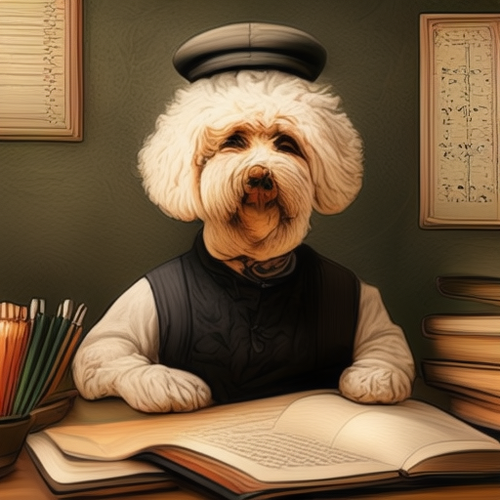} &
\includegraphics[scale=0.161,valign=t]{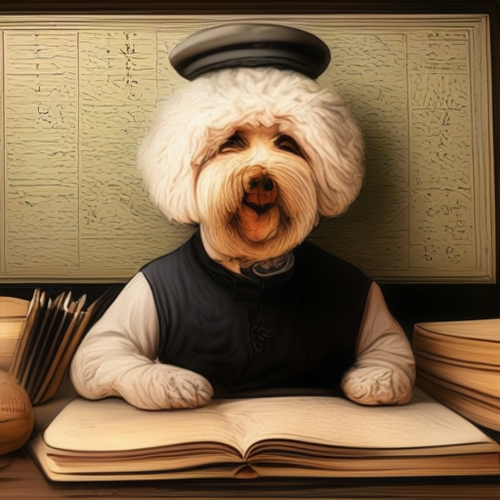} &
\includegraphics[scale=0.161,valign=t]{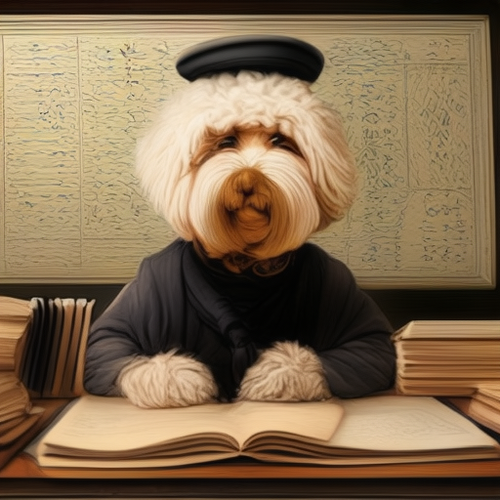} &
\includegraphics[scale=0.161,valign=t]{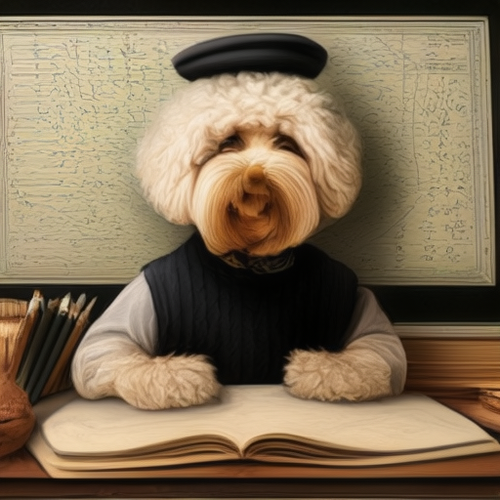} &
\includegraphics[scale=0.161,valign=t]{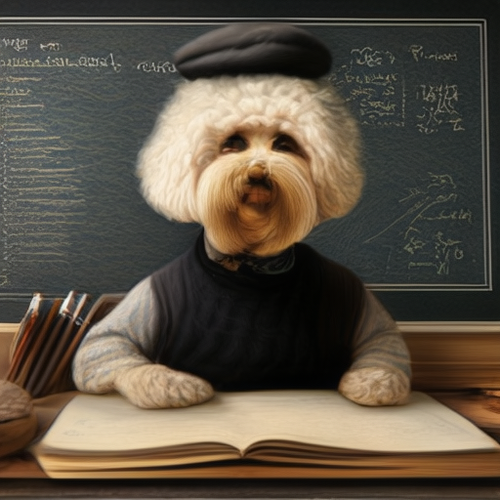} &
\includegraphics[scale=0.161,valign=t]{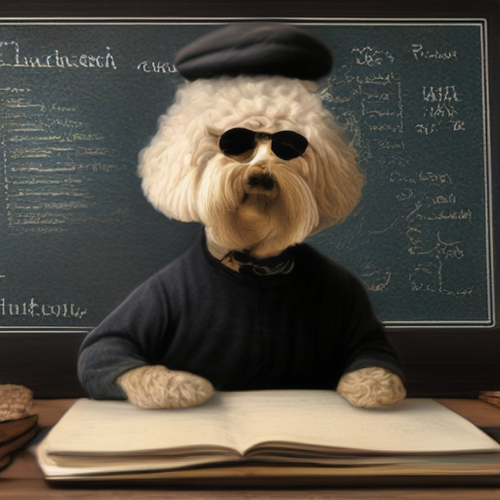} \\

\multicolumn{6}{c}{\emph{a cream-colored labradoodle wearing glasses and black beret teaching calculus at a blackboard}}  \\ 
\bottomrule
\end{tabular}
\endgroup
    \caption{}
\end{figure*}

\begin{figure*}[ht]
\centering
\begingroup

\setlength{\tabcolsep}{0pt} %
\renewcommand{\arraystretch}{-100} %
\begin{tabular}{ 0{p{0.166\textwidth}}  0{p{0.166\textwidth}}  0{p{0.166\textwidth}} 0{p{0.166\textwidth}} 0{p{0.166\textwidth}} 0{p{0.166\textwidth}} }\toprule

\includegraphics[scale=0.161,valign=t]{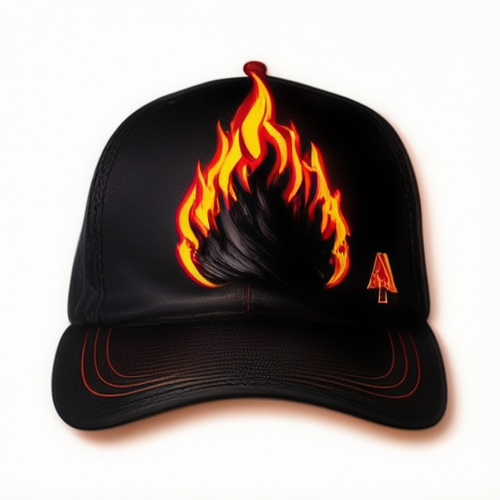}\\
\includegraphics[scale=0.161,valign=t]{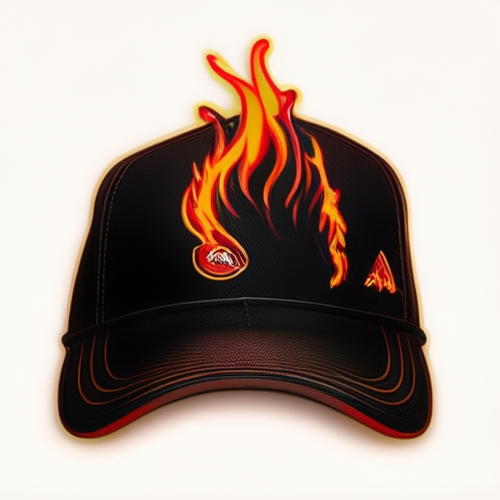} &
\includegraphics[scale=0.161,valign=t]{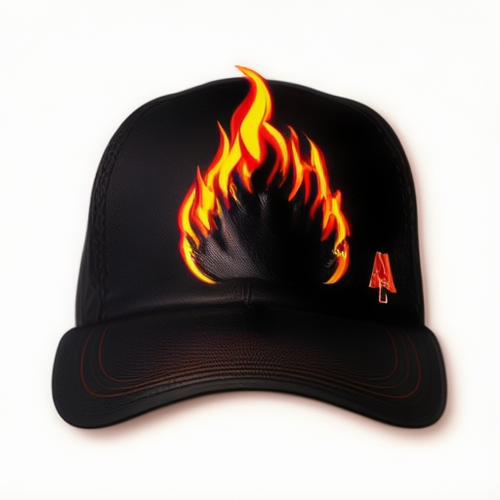}\\
\includegraphics[scale=0.161,valign=t]{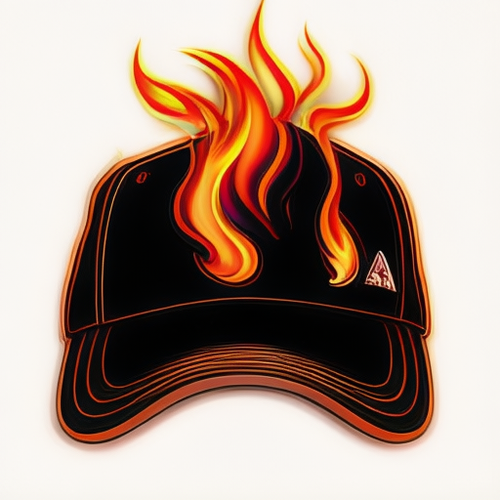} &
\includegraphics[scale=0.161,valign=t]{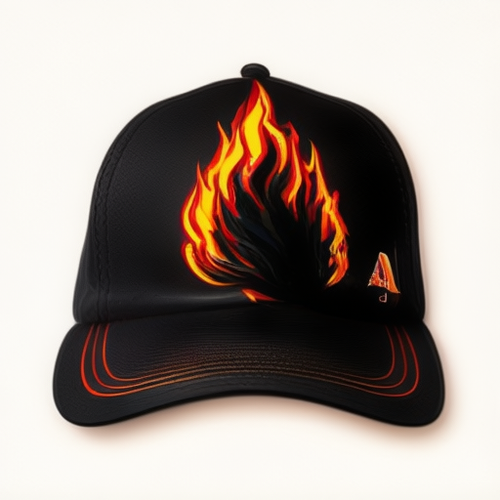} &
\includegraphics[scale=0.161,valign=t]{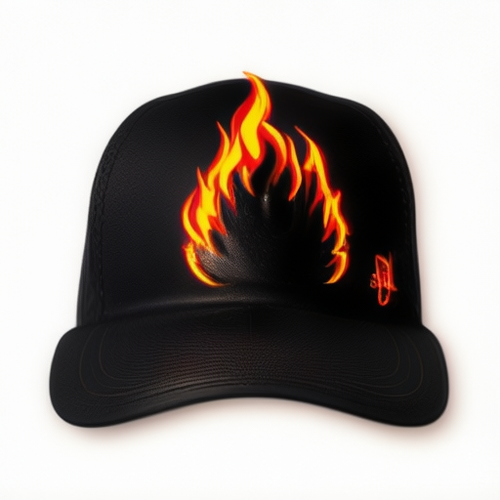}\\
\includegraphics[scale=0.161,valign=t]{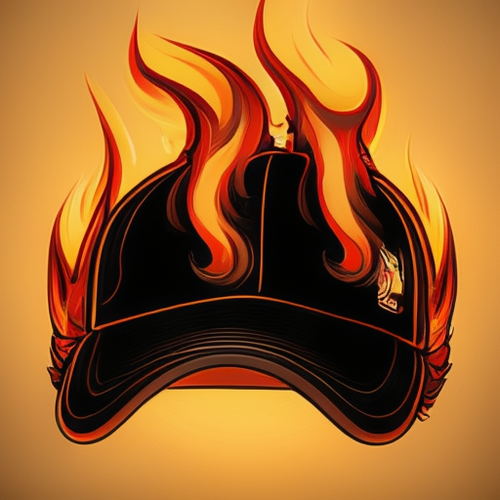} &
\includegraphics[scale=0.161,valign=t]{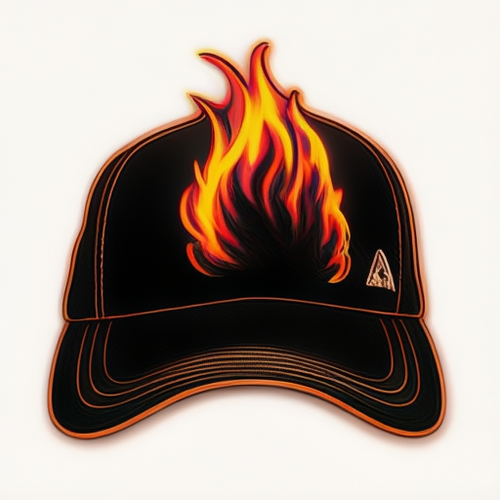} &
\includegraphics[scale=0.161,valign=t]{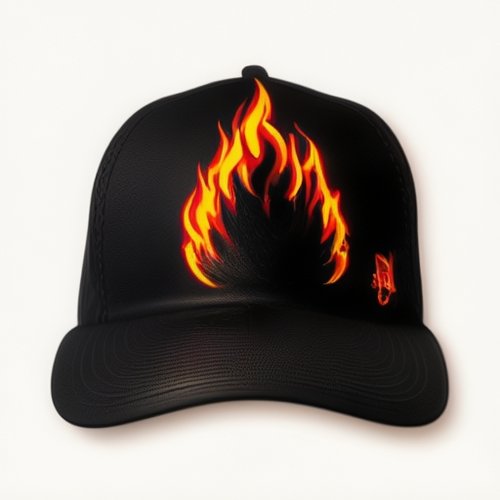} &
\includegraphics[scale=0.161,valign=t]{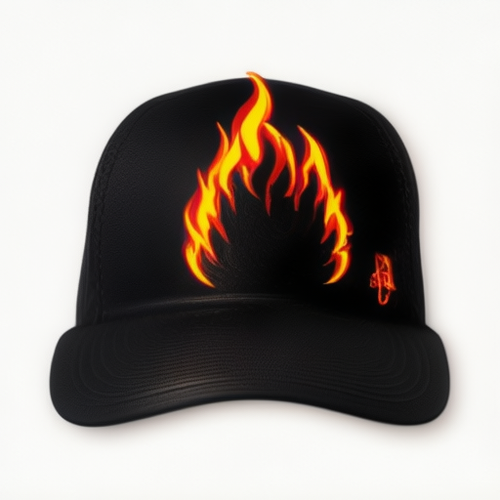} \\
\includegraphics[scale=0.161,valign=t]{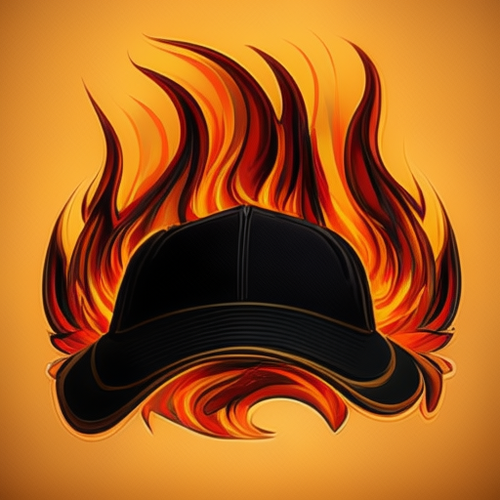} &
\includegraphics[scale=0.161,valign=t]{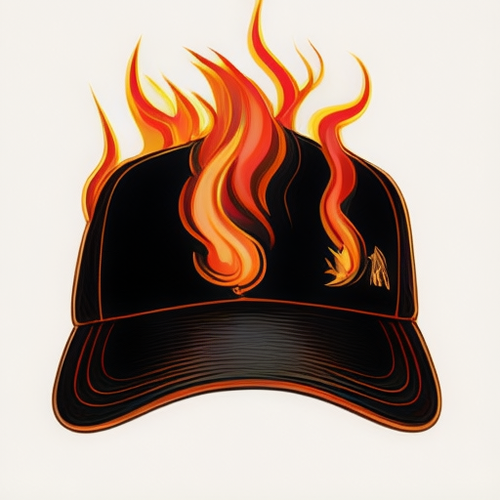} &
\includegraphics[scale=0.161,valign=t]{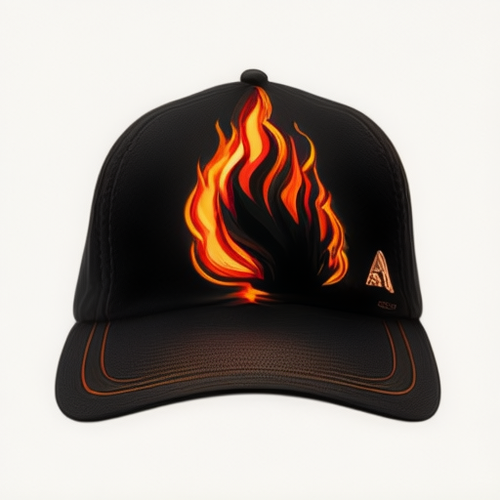} &
\includegraphics[scale=0.161,valign=t]{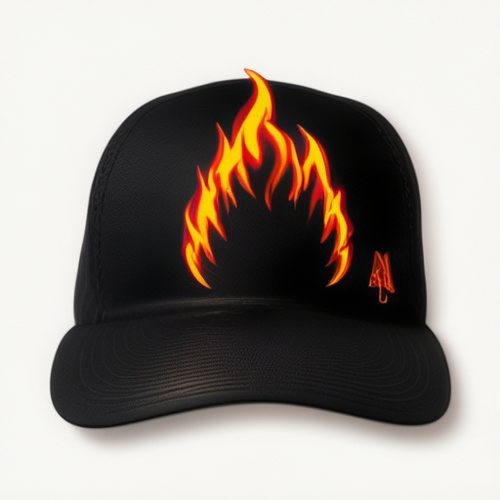} &
\includegraphics[scale=0.161,valign=t]{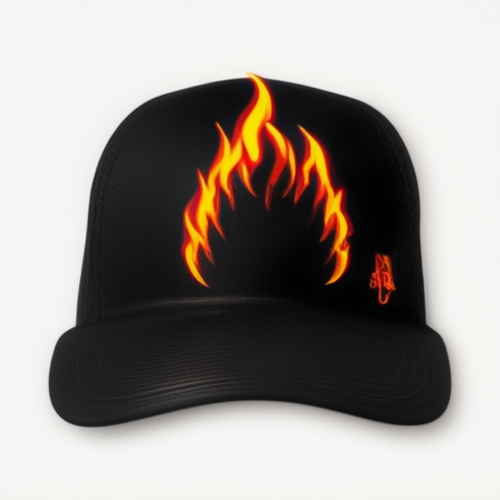} \\
\includegraphics[scale=0.161,valign=t]{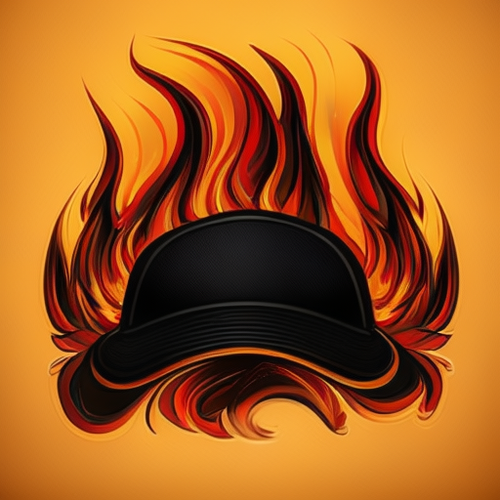} &
\includegraphics[scale=0.161,valign=t]{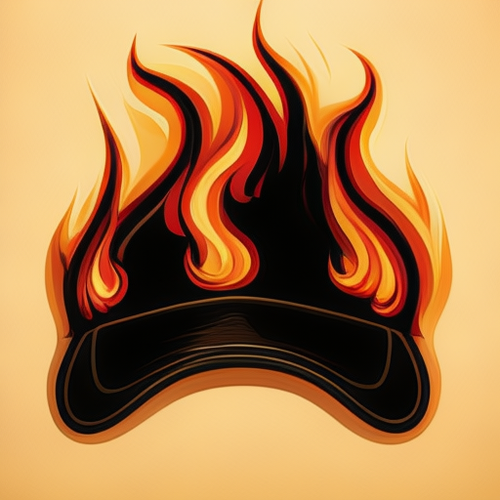} &
\includegraphics[scale=0.161,valign=t]{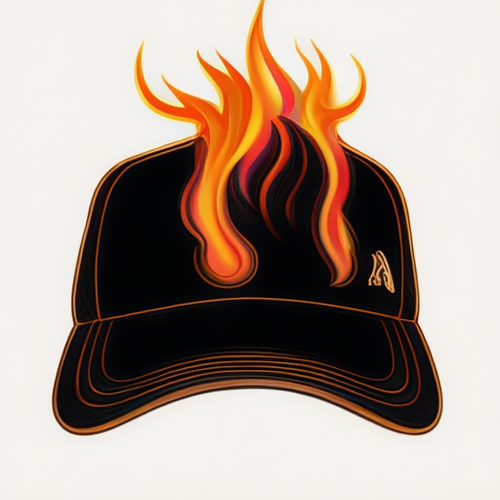} &
\includegraphics[scale=0.161,valign=t]{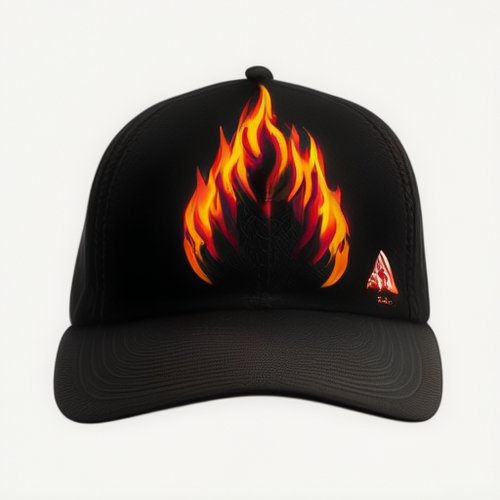} &
\includegraphics[scale=0.161,valign=t]{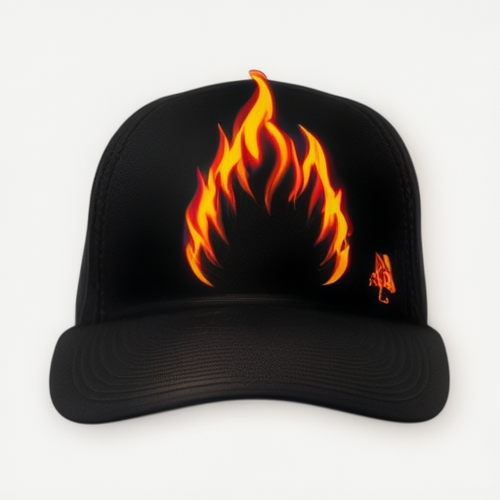} &
\includegraphics[scale=0.161,valign=t]{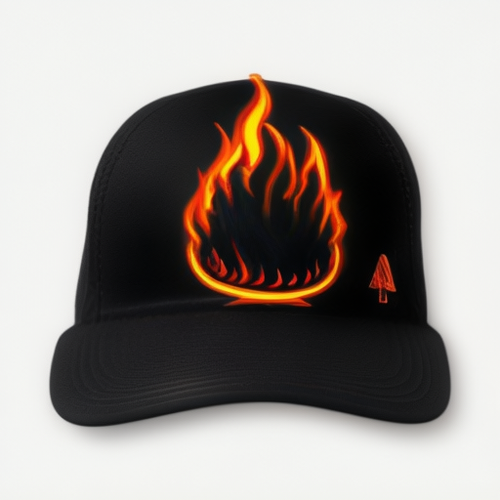} \\

\multicolumn{6}{c}{\emph{a black baseball hat with a flame decal on it}}  \\ 
\bottomrule
\end{tabular}
\endgroup
    \caption{}
\end{figure*}

\begin{figure*}[ht]
\centering
\begingroup

\setlength{\tabcolsep}{0pt} %
\renewcommand{\arraystretch}{-100} %
\begin{tabular}{ 0{p{0.166\textwidth}}  0{p{0.166\textwidth}}  0{p{0.166\textwidth}} 0{p{0.166\textwidth}} 0{p{0.166\textwidth}} 0{p{0.166\textwidth}} }\toprule

\includegraphics[scale=0.161,valign=t]{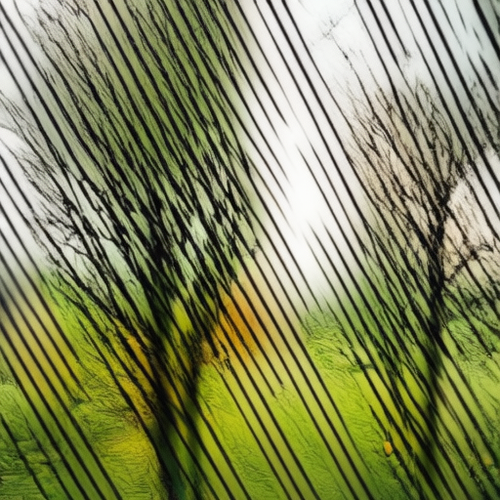}\\
\includegraphics[scale=0.161,valign=t]{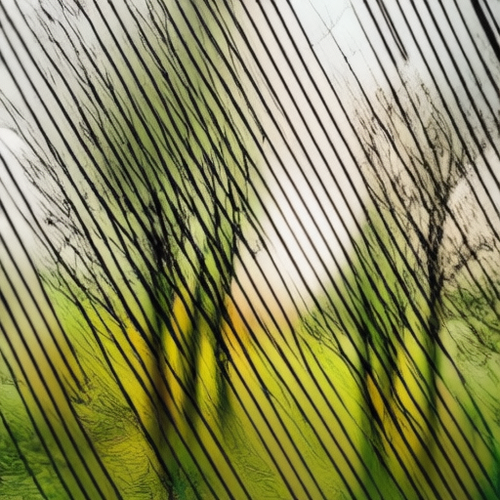} &
\includegraphics[scale=0.161,valign=t]{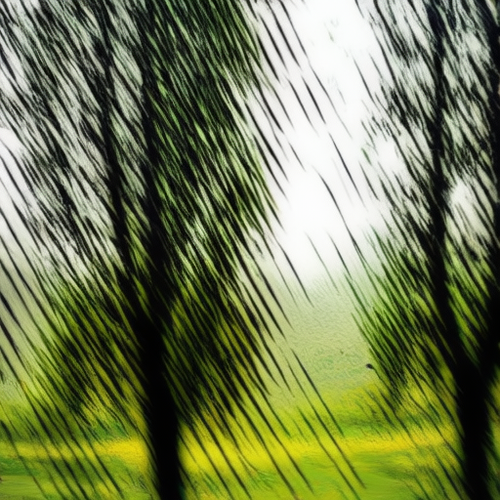}\\
\includegraphics[scale=0.161,valign=t]{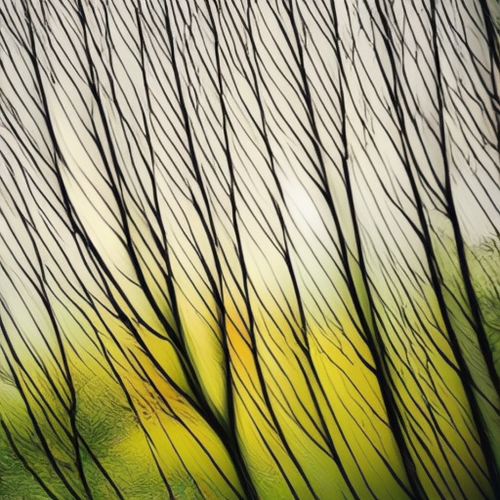} &
\includegraphics[scale=0.161,valign=t]{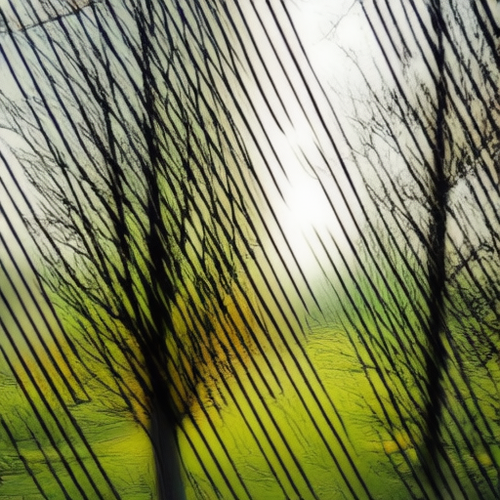} &
\includegraphics[scale=0.161,valign=t]{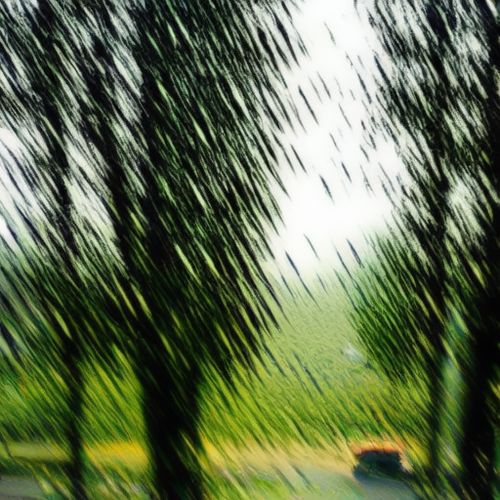}\\
\includegraphics[scale=0.161,valign=t]{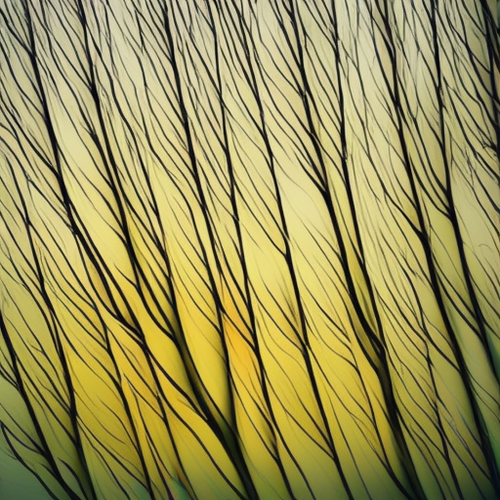} &
\includegraphics[scale=0.161,valign=t]{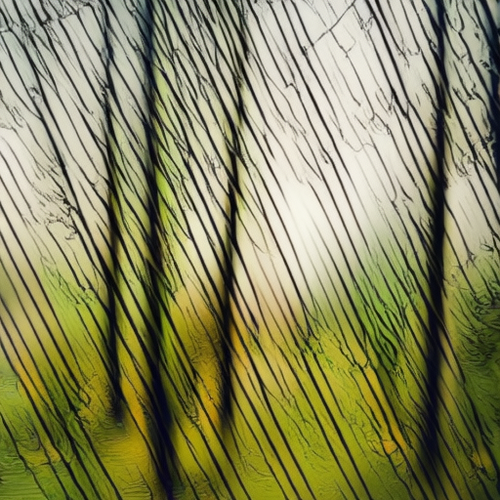} &
\includegraphics[scale=0.161,valign=t]{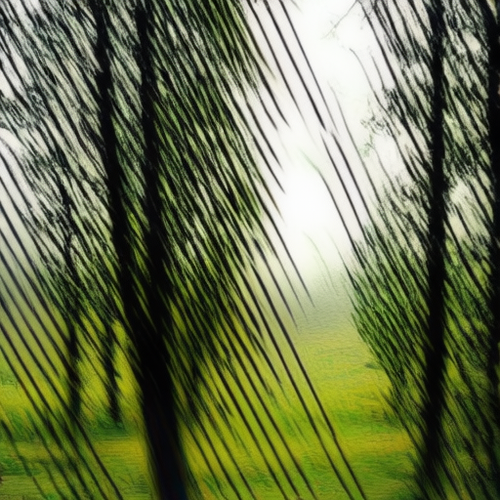} &
\includegraphics[scale=0.161,valign=t]{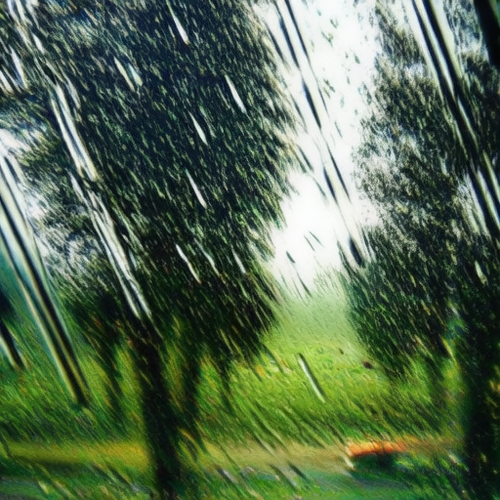} \\
\includegraphics[scale=0.161,valign=t]{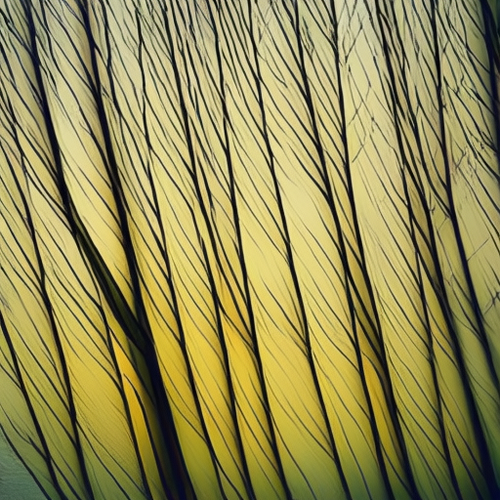} &
\includegraphics[scale=0.161,valign=t]{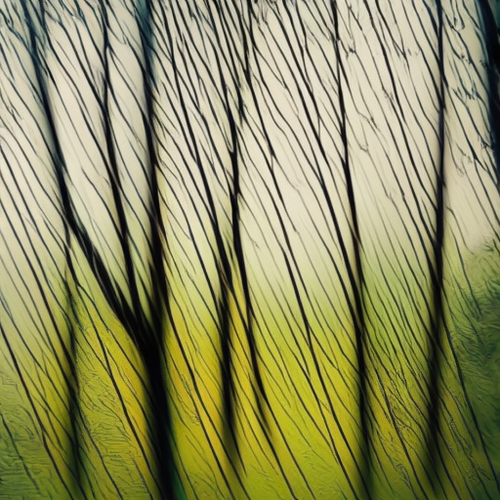} &
\includegraphics[scale=0.161,valign=t]{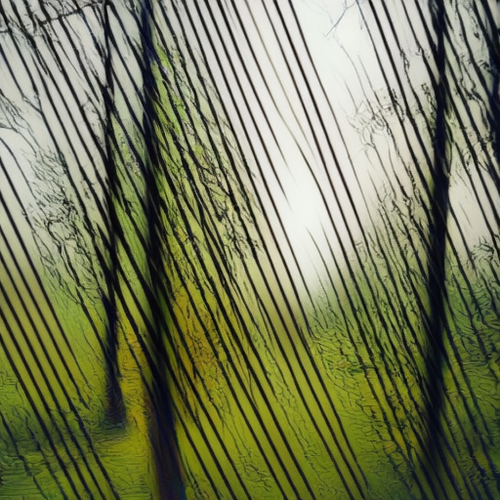} &
\includegraphics[scale=0.161,valign=t]{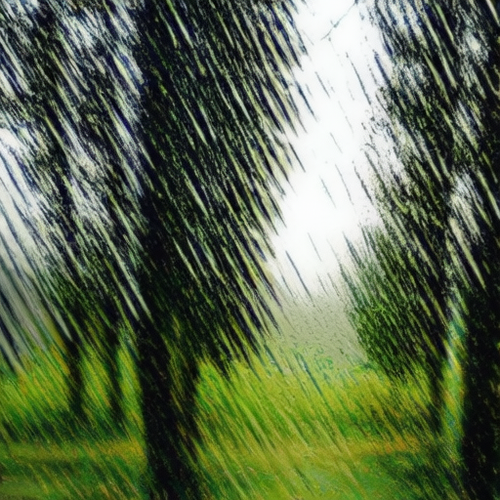} &
\includegraphics[scale=0.161,valign=t]{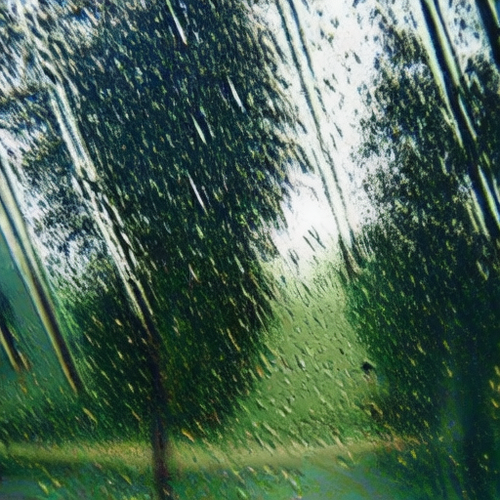} \\
\includegraphics[scale=0.161,valign=t]{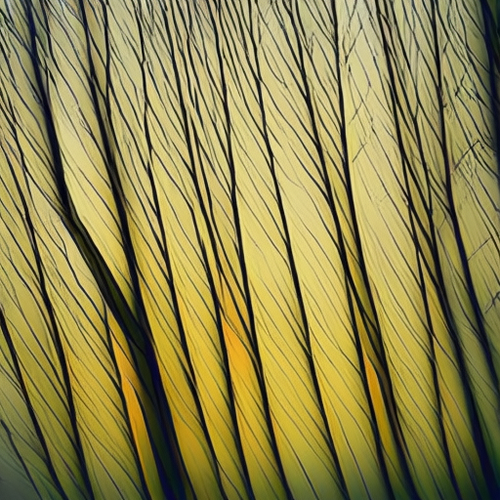} &
\includegraphics[scale=0.161,valign=t]{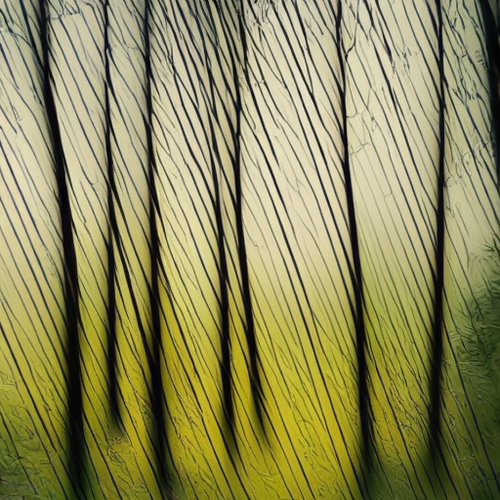} &
\includegraphics[scale=0.161,valign=t]{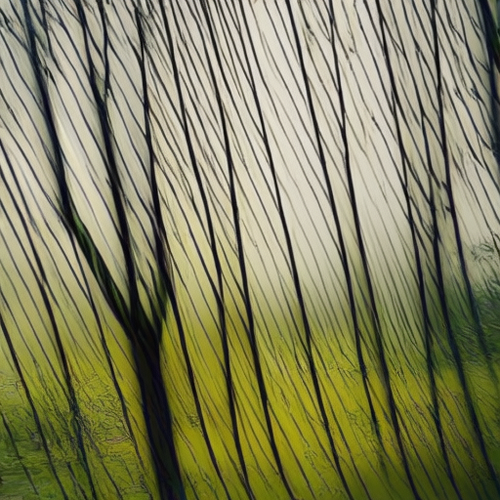} &
\includegraphics[scale=0.161,valign=t]{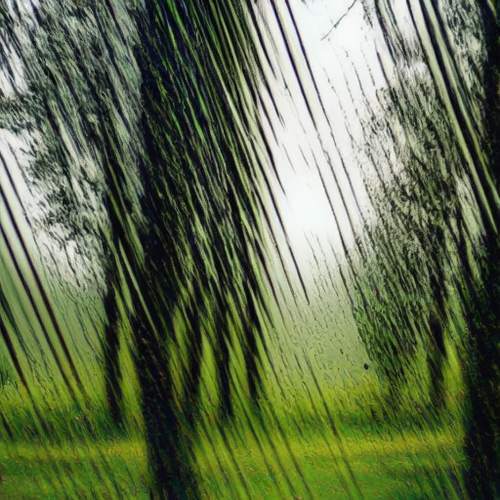} &
\includegraphics[scale=0.161,valign=t]{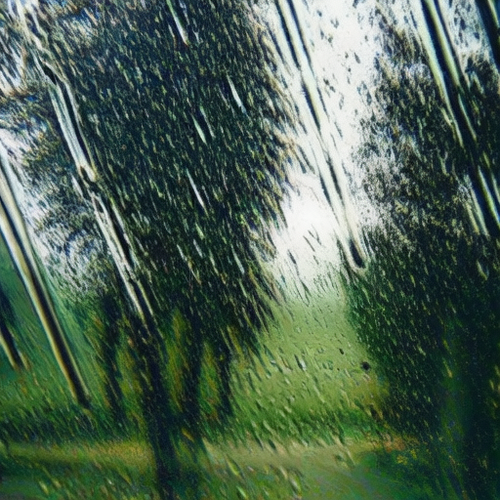} &
\includegraphics[scale=0.161,valign=t]{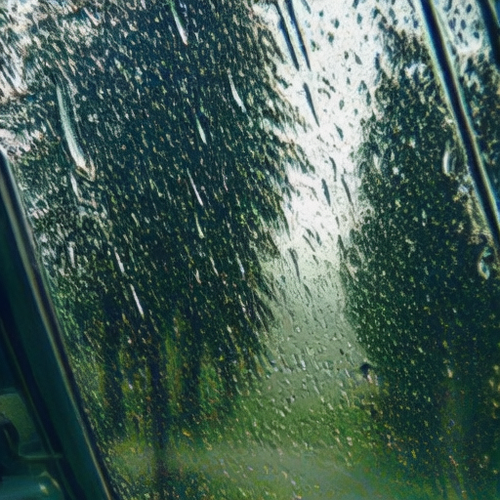} \\

\multicolumn{6}{c}{\emph{trees seen through a car window on a rainy day}}  \\ 
\bottomrule
\end{tabular}
\endgroup
    \caption{}
\end{figure*}

\end{document}